\documentclass[sigconf,natbib=true,screen]{acmart}
\usepackage{xspace,balance,tabularx,multirow}
\usepackage{flushend}
\usepackage{tikz}
\usepackage{pgfplots}
\pgfplotsset{compat=1.16}
\usepackage{tabularx}
\usetikzlibrary{patterns}
\usepackage{subfig}
\usepackage[ruled, vlined, linesnumbered]{algorithm2e}
\usepackage{xcolor}
\usepackage{colortbl}
\usepackage{bbold}
\SetKwComment{Comment}{$\triangleright$\ }{}
\usepackage{enumitem}
\usepackage{tablefootnote}
\usepackage{upgreek,textgreek}
\usepackage{pifont}
\usepackage[noabbrev]{cleveref}
\usepackage{titlecaps}
\usepackage{lipsum}
\usepackage{graphicx}

\pgfplotsset{every tick label/.append style={font=\tiny}}

\newlength{\starsize}
\newlength{\starspread}
\tikzset{starsize/.code={\setlength{\starsize}{#1}},
         starspread/.code={\setlength{\starspread}{#1}}}
\tikzset{starsize=1mm,
         starspread=3mm}
\pgfdeclarepatternformonly[\starspread,\starsize]
  {my fivepointed stars}
  {\pgfpointorigin}
  {\pgfqpoint{\starspread}{\starspread}}
  {\pgfqpoint{\starspread}{\starspread}}
  {
   \pgftransformshift{\pgfqpoint{\starsize}{\starsize}}
   \pgfpathmoveto{\pgfqpointpolar{18}{\starsize}}
   \pgfpathlineto{\pgfqpointpolar{162}{\starsize}}
   \pgfpathlineto{\pgfqpointpolar{306}{\starsize}}
   \pgfpathlineto{\pgfqpointpolar{90}{\starsize}}
   \pgfpathlineto{\pgfqpointpolar{234}{\starsize}}
   \pgfpathclose%
   \pgfusepath{fill}
  }

\newcommand{\argtopk}[1]{\underset{#1}{\operatorname{arg}\,\operatorname{topk}}\;}

\makeatletter
\newcommand*\bigcdot{\mathpalette\bigcdot@{.5}}
\newcommand*\bigcdot@[2]{\mathbin{\vcenter{\hbox{\scalebox{#2}{$\m@th#1\bullet$}}}}}
\makeatother

\newcommand{\stitle}[1]{\vspace*{0.5em}\noindent{\bf #1.\/}}

\newcommand{\V}{\mathcal{V}\xspace}
\newcommand{\VTR}{\mathcal{V}_{\text{tr}}\xspace}
\newcommand{\YTR}{\mathcal{Y}_{\text{tr}}\xspace}
\newcommand{\G}{\mathcal{G}\xspace}
\newcommand{\N}{\mathcal{N}\xspace}
\newcommand{\EDG}{\mathcal{E}\xspace}
\newcommand{\T}{\mathcal{T}\xspace}
\newcommand{\C}{\mathcal{C}\xspace}
\newcommand{\A}{\mathcal{A}\xspace}
\newcommand{\HH}{\mathcal{H}\xspace}
\newcommand{\Sset}{\mathcal{S}\xspace}

\newcommand{\VCS}{\mathcal{V}_{\text{ct}}\xspace}
\newcommand{\VUS}{\mathcal{V}_{\text{ut}}\xspace}

\newcommand{\AM}{\boldsymbol{A}\xspace}
\newcommand{\DM}{\boldsymbol{D}\xspace}
\newcommand{\IM}{\boldsymbol{I}\xspace}
\newcommand{\SM}{\boldsymbol{S}\xspace}

\newcommand{\CM}{\boldsymbol{C}\xspace}
\newcommand{\YM}{\boldsymbol{Y}\xspace}
\newcommand{\XM}{\boldsymbol{X}\xspace}
\newcommand{\LM}{\boldsymbol{L}\xspace}
\newcommand{\UM}{\boldsymbol{U}\xspace}

\newcommand{\HM}{\boldsymbol{H}\xspace}
\newcommand{\ZM}{\boldsymbol{Z}\xspace}

\newcommand{\NAM}{\boldsymbol{\tilde{A}}\xspace}

\newcommand{\BPA}{B_{\text{ini}}\xspace}
\newcommand{\BLC}{B_{\text{ref}}\xspace}

\newcommand{\algo}{\texttt{Locle}\xspace}

\newcommand{\eat}[1]{}

\newenvironment{customlegend}[1][]{%
    \begingroup
    \csname pgfplots@init@cleared@structures\endcsname
    \pgfplotsset{#1}%
}{%
    \csname pgfplots@createlegend\endcsname
    \endgroup
}%

\def\addlegendimage{\csname pgfplots@addlegendimage\endcsname}

\makeatletter
\newcommand\footnoteref[1]{\protected@xdef\@thefnmark{\ref{#1}}\@footnotemark}
\makeatother

\let\oldnl\nl
\newcommand{\nonl}{\renewcommand{\nl}{\let\nl\oldnl}}

\DeclareMathOperator{\Tr}{\textsf{trace}}

\SetKwComment{Comment}{/* }{ */}

\makeatletter 
\g@addto@macro{\@algocf@init}{\SetKwInOut{Parameter}{Parameters}} 
\makeatother

\definecolor{myred}{HTML}{fd7f6f}
\definecolor{myred_new}{HTML}{D8D8D8}
\definecolor{myred_new2}{HTML}{D7191C}
\definecolor{myblue}{HTML}{7eb0d5}
\definecolor{mygreen}{HTML}{b2e061}
\definecolor{mypurple}{HTML}{bd7ebe}
\definecolor{myorange}{HTML}{ffb55a}
\definecolor{myyellow}{HTML}{ffee65}
\definecolor{mypurple2}{HTML}{beb9db}
\definecolor{mypink}{HTML}{fdcce5}
\definecolor{mycyan}{HTML}{8bd3c7}

\definecolor{myblue2}{HTML}{115f9a}
\definecolor{myred2}{HTML}{c23728}

\AtBeginDocument{%
  \providecommand\BibTeX{{%
    \normalfont B\kern-0.5em{\scshape i\kern-0.25em b}\kern-0.8em\TeX}}}

\copyrightyear{2025}
\acmYear{2025}
\setcopyright{cc}
\setcctype{by}
\acmConference[SIGIR '25]{Proceedings of the 48th International ACM SIGIR Conference on Research and Development in Information Retrieval}{July 13--18, 2025}{Padua, Italy}
\acmBooktitle{Proceedings of the 48th International ACM SIGIR Conference on Research and Development in Information Retrieval (SIGIR '25), July 13--18, 2025, Padua, Italy}\acmDOI{10.1145/3726302.3730021}
\acmISBN{979-8-4007-1592-1/2025/07}

\acmSubmissionID{1035}

\begin{document}


\title{Leveraging Large Language Models for Effective Label-free Node Classification in Text-Attributed Graphs}
\subtitle{Technical Report}
\author{Taiyan Zhang}
\authornote{Work done while at HKBU.}
\authornote{Both authors contributed equally to the paper.}
\affiliation{%
  \institution{ShanghaiTech University
  \\
  Hong Kong Baptist University
  }
  \country{}
}
\email{zhangty2022@shanghaitech.edu.cn}

\author{Renchi Yang}
\authornotemark[2]
\affiliation{%
  \institution{Hong Kong Baptist University}
  \country{}
}
\email{renchi@hkbu.edu.hk}

\author{Yurui Lai}
\affiliation{%
  \institution{Hong Kong Baptist University}
  \country{}
}
\email{csyrlai@comp.hkbu.edu.hk}

\author{Mingyu Yan}
\authornote{Corresponding Author}
\affiliation{%
   \institution{
    Institute of Computing Technology, Chinese Academy of Sciences 
    }
  \country{}
}
\email{yanmingyu@ict.ac.cn}

\author{Xiaochun Ye}
\affiliation{%
   \institution{
    Institute of Computing Technology, Chinese Academy of Sciences 
    }
  \country{}
}
\email{yexiaochun@ict.ac.cn}

\author{Dongrui Fan}
\affiliation{%
   \institution{
    Institute of Computing Technology, Chinese Academy of Sciences 
    }
  \country{}
}
\email{fandr@ict.ac.cn}


\begin{abstract}
Graph neural networks (GNNs) have become the preferred models for node classification in graph data due to their robust capabilities in integrating graph structures and attributes. However, these models heavily depend on a substantial amount of high-quality labeled data for training, which is often costly to obtain. With the rise of large language models (LLMs), a promising approach is to utilize their exceptional zero-shot capabilities and extensive knowledge for node labeling. Despite encouraging results, this approach either requires numerous queries to LLMs or suffers from reduced performance due to noisy labels generated by LLMs. To address these challenges, we introduce \textbf{Locle}, an active self-training framework that does \underline{\textbf{L}}abel-free n\underline{\textbf{O}}de \underline{\textbf{C}}lassification with \underline{\textbf{L}}LMs cost-\underline{\textbf{E}}ffectively. Locle iteratively identifies small sets of "critical" samples using GNNs and extracts informative pseudo-labels for them with both LLMs and GNNs, serving as additional supervision signals to enhance model training. Specifically, Locle comprises three key components: (i) an effective active node selection strategy for initial annotations; (ii) a careful sample selection scheme to identify "critical" nodes based on label disharmonicity and entropy; and (iii) a label refinement module that combines LLMs and GNNs with a rewired topology. Extensive experiments on five benchmark text-attributed graph datasets demonstrate that Locle significantly outperforms state-of-the-art methods under the same query budget to LLMs in terms of label-free node classification. Notably, on the DBLP dataset with 14.3k nodes, Locle achieves an 8.08\% improvement in accuracy over the state-of-the-art at a cost of less than one cent. Our code is available at \url{https://github.com/HKBU-LAGAS/Locle}.

\end{abstract}

\ccsdesc[500]{Computing methodologies~Neural networks}
\ccsdesc[500]{Information systems~Graph-based data models}

\keywords{Graph Neural Network, Large Language Models, Label-free Node Classification}
\maketitle

\section{Introduction}\label{sec:intro}

{\em Text-attributed graphs} (TAGs)~\cite{chang2009relational} are an expressive data model used to represent textual entities and their complex interconnections. Such data structures are prevalent in real-world scenarios, including social networks, hyperlink graphs of web pages, transaction networks, etc., wherein nodes are endowed with user profiles, web page contents, or product descriptions.
Node classification is a fundamental task over TAGs, which aims to classify the nodes in the graph into a number of predefined categories based on the graph structures and textual contents. This task has emerged as a critical area of research in Information Retrieval (IR)~\cite{fu2021sdg}. For instance, in content-sharing social networks like Flickr, content classification enables topical filtering and tag-based retrieval of multimedia items~\cite{seah2018killing}. In product search contexts, the retrieval of advertisement item tags can be framed as a conventional classification task~\cite{mao2020item}. Similarly, in e-commerce query graphs, classifying query intent enhances result ranking by aligning it with the intended product category~\cite{fang2012confidence}.

In the past decade, {\em graph neural networks} (GNNs)~\cite{kipf2016semi,gasteiger2018predict,xu2018representation,chen2020simple,wu2019simplifying,huang2023node} have become the dominant models for node classification, by virtue of their capabilities to capture complex dependencies and patterns in the graph and the interplay between connectivity and attributes.
However, the efficacy of such models largely hinges on the availability of adequate node labels for training. 
In real life, high-quality labels are hard to acquire due to the need for expert knowledge, significant human efforts, and potential biases in annotation, particularly for large-scale graphs comprising millions of nodes and a sheer volume of textual data~\cite{dai2021nrgnnlearninglabelnoiseresistant, sheng2008get}.

In light of the superb comprehension and reasoning abilities of {\em large language models} (LLMs) in dealing with textual data, LLMs have been employed as a powerful tool for analyzing TAGs. 
As manifested in ~\cite{chen2024exploring,heharnessing}, LLMs have exhibited impressive node classification performance in TAGs under zero-shot or few-shot settings. 
However, compared to GNNs, this methodology falls short of exploiting the graph structures and requires substantial queries/calls to LLMs, and hence is not suitable for classifying nodes in the entire graph.
Instead, a recent study~\cite{chen2023label} has made an attempt towards utilizing LLMs as annotators for node labeling to facilitate the zero-shot node classification over TAGs.
In this pipeline, i.e., \texttt{LLM-GNN}, LLMs work as a front-mounted step to create labels for a small number of selected {\em active} nodes as training data, and subsequently, a GNN model is trained for classification with the labeled data.

\begin{figure}[!t]
\centering
\begin{small}
\begin{tikzpicture}
    \begin{customlegend}
    [legend columns=2,
        legend entries={\texttt{LLM-GNN} (ground-truth), \texttt{LLM-GNN}},
        legend style={at={(0.45,1.35)},anchor=north,draw=none,font=\small,column sep=0.2cm}]
    \addlegendimage{line width=0.4mm,mark size=3pt,mark=square,color=teal}
    \addlegendimage{line width=0.4mm,mark size=3pt,mark=star,color=orange}
    \end{customlegend}
\end{tikzpicture}
\\[-\lineskip]
\vspace{-3mm}
\subfloat[\em Cora]{
\begin{tikzpicture}[scale=1,every mark/.append style={mark size=2pt}]
    \begin{axis}[
        height=\columnwidth/2.5,
        width=\columnwidth/1.9,
        ylabel={\it Accuracy},
        xmin=0.5, xmax=13.5,
        ymin=0.6, ymax=0.9,
        xtick={1,3,5,7,9,11,13},
        ytick={0.6,0.7,0.8,0.9},
        xticklabel style = {font=\footnotesize},
        yticklabel style = {font=\small},
        xticklabels={70,105,140,175,210,245,280},
        yticklabels={0.6,0.7,0.8,0.9},
        every axis y label/.style={font=\small,at={(current axis.north west)},right=5mm,above=0mm},
        legend style={fill=none,font=\small,at={(0.02,0.99)},anchor=north west,draw=none},
    ]
    \addplot[line width=0.4mm, mark=square,color=teal]  
        plot coordinates {
(1,	0.7844	)
(3,	0.8302	)
(5,	0.8208	)
(7,	0.8254	)
(9,	0.8304	)
(11,0.8236	)
(13, 0.8)
};

    \addplot[line width=0.4mm, mark=star,color=orange]  
        plot coordinates {
(1,	0.7005	)
(3,	0.7514	)
(5,	0.7475	)
(7,	0.7273	)
(9, 0.7245)
(11, 0.7250	)
(13, 0.7147)
    };

    \end{axis}
\end{tikzpicture}\hspace{4mm}\label{fig:budget-cora}%
}
\subfloat[\em CiteSeer]{
\begin{tikzpicture}[scale=1,every mark/.append style={mark size=2pt}]
    \begin{axis}[
        height=\columnwidth/2.5,
        width=\columnwidth/1.9,
        ylabel={\it Accuracy},
        xmin=0.5, xmax=11.5,
        ymin=0.5, ymax=0.8,
        xtick={1,3,5,7,9,11},
        ytick={0.5,0.6,0.7,0.8},
        xticklabel style = {font=\small},
        yticklabel style = {font=\small},
        xticklabels={90,120 ,150, 180, 210,240 },
        yticklabels={0.5,0.6,0.7,0.8},
        every axis y label/.style={font=\small,at={(current axis.north west)},right=5mm,above=0mm},
        legend style={fill=none,font=\small,at={(0.02,0.99)},anchor=north west,draw=none},
    ]
    \addplot[line width=0.4mm, mark=square,color=teal]  
        plot coordinates {
(1,	0.6886	)
(3,	0.7071	)
(5,	0.7129	)
(7,	0.7252	)
(9,	0.7375	)
(11,	0.7493	)
    };

   \addplot[line width=0.4mm, mark=star,color=orange]  
        plot coordinates {
(1,	0.6396 )
(3,	0.6123	)
(5,	0.6063	)
(7,	0.6200	)
(9,	0.6306	)
(11,	0.6246	)
};

    \end{axis}
\end{tikzpicture}\hspace{0mm}\label{fig:budget-citeseer}%
}
\end{small}
 \vspace{-3mm}
\caption{Varying \#labeled nodes.} \label{fig:example}
\vspace{-4ex}
\end{figure}
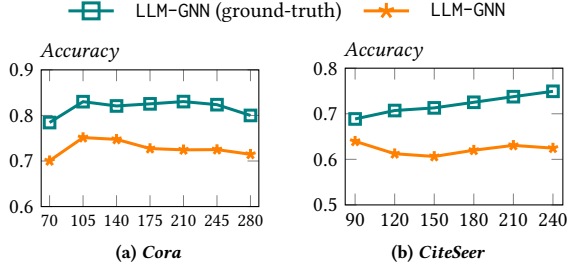

Despite its empirical effectiveness as reported in~\cite{chen2023label}, \texttt{LLM-GNN} is inherently defective due to its cascaded workflow that strongly relies on the active node selection and output quality of LLMs, which could be inaccurate/noisy, and in turn, leads to sub-optimal node classification performance. 
To illustrate, Figure~\ref{fig:example} depicts the empirical performance attained by \texttt{LLM-GNN} and its variant whose active nodes are annotated with the ground-truth labels, dubbed as \texttt{LLM-GNN} (ground-truth), on two real TAGs (see Section~\ref{sec:exp}), when increasing the labeled sample size.
The first observation we can make from Figure~\ref{fig:example} is that there is a notable performance gap (around $10\%$) between \texttt{LLM-GNN} and \texttt{LLM-GNN} (ground-truth). Second, as the labeled sample size, i.e., query budget for LLMs, is increased, \texttt{LLM-GNN} undergoes performance stagnation or even degradation on both datasets. These observations indicate that the label noise introduced by LLMs severely undermines the performance of GNNs. Moreover, even for \texttt{LLM-GNN} (ground-truth), we can also observe a drop of performance on {\em Cora} when training with more labeled data, which reveals the limitations of \texttt{LLM-GNN} in selecting representative active samples for labeling as well as vanilla GNN models.


To address the above-said problems, we present \algo (\underline{\textbf{L}}abel-free n\underline{\textbf{O}}de \underline{\textbf{C}}lassification with \underline{\textbf{L}}LMs cost-\underline{\textbf{E}}ffectively. ), a new framework that integrates LLMs into GNNs for cost-effective zero-shot node classification.
To achieve this goal, the basic idea of \algo is to actively capitalize on the rich structural and attribute semantics underlying the TAGs with GNNs and LLMs for active node selection, node annotation, and label correction, rather than solely based on LLMs.
More concretely, \algo proceeds in two stages that both combine LLMs and GNNs.
Given a total query budget $B$ to access LLMs, the first stage in \algo pinpoints $\varepsilon\cdot B$ ($0<\varepsilon<1$) active nodes for annotation by LLMs through the {\em subspace clustering}~\cite{vidal2011subspace} based on GNN-based node representations, which enables the accurate selection of representative node samples with consideration of the inherent structures of the input graph and attribute data.
The second stage resorts to a multi-round self-training pipeline, in which GNNs focus on identifying a set of ``informative'' samples with high-confidence and low-confidence labels using our proposed {\em label entropy} and {\em label disharmonicity} metrics, while the LLM is harnessed to generate more reliable labels for these uncertain samples with the remaining $(1-\varepsilon)\cdot B$ query budget. To reduce the label noise from LLMs, we further propose a {\em Dirichlet Energy}-based graph rewiring strategy to minimize the adverse effects of noisy or missing links in the original graphs, 
whereby \algo can infer new label predictions to refine the LLM-based annotations.
The pseudo-labels for such informative samples will be further used as additional supervision signals to facilitate the subsequent model training.

To summarize, our contributions in this paper are as follows:
\begin{itemize}[leftmargin=*]
\item We introduce a novel multi-round self-training framework that enables the cost-effective integration of LLMs and GNNs for improved label-free node classification. 
\item We propose an effective active node selection scheme for node annotations with LLMs. 
\item We design a judicious strategy for the selection of informative samples and a graph rewiring-based label refinement to create more reliable labeled data for model training.
\item Extensive experiments comparing \algo against a total of 19 baselines over five real TAGs showcase that \algo can achieve a consistent and remarkable improvement of at least $5\%$ in zero-shot classification accuracy compared to the state of the art in most cases.
\end{itemize}

\section{Related Work}
This section reviews existing studies germane to our work.

\subsection{Zero-shot Node Classification}
Zero-shot node classification aims to train a model on a set of known categories and generalize it to unseen categories. GraphCEN~\cite{ju2023zero} introduces a two-level contrastive learning approach to jointly learn node embeddings and class assignments in an end-to-end fashion, effectively enabling the transfer of knowledge to unseen classes. Similarly, TAG-Z~\cite{li2023prompt} leverages prompts alongside graph topology to generate preliminary logits, which can be directly applied to zero-shot node classification tasks. DGPN~\cite{wang2021zero} facilitates zero-shot knowledge transfer by utilizing class semantic descriptions to transfer knowledge from seen to unseen categories, a process analogous to meta-learning. Additionally, methods such as BART~\cite{lewis2019bartdenoisingsequencetosequencepretraining} can also be adapted for node classification tasks. However, it is important to note that our approach differs slightly from traditional zero-shot node classification, as we do not require any initial training data.

\subsection{Node Classification on Text-Attributed Graphs}

Text-attributed graphs combine two modalities: the textual content within documents and the graph structure that connects these documents~\cite{chang2009relational}. On the one hand, Graph Neural Networks (GNNs)~\cite{hamilton2017inductive} effectively generate document embeddings by integrating both vertex attributes and graph connectivity. However, most existing GNN-based models treat textual content as general attributes without specifically addressing the unique properties of language data. Consequently, they fail to capture the rich semantic structures and nuanced language representations embedded in text corpora.

On the other hand, pre-trained language models (PLMs)~\cite{vaswani2017attention} and large language models (LLMs) excel at learning contextualized language representations and generating document embeddings. However, these models typically focus on individual documents and do not consider the graph connectivity between documents, such as citations or hyperlinks. This connectivity often encodes topic similarity, and by modeling it, one can propagate semantic information across connected documents.

To address these challenges, recent approaches have proposed text-attributed graph representation learning, which combines GNNs with PLMs and LLMs into unified frameworks for learning document embeddings that preserve both contextualized textual semantics and graph connectivity. For instance, Graphformers~\cite{zhang2020graph} iteratively integrate text encoding with graph aggregation, enabling each node’s semantics to be understood from a global perspective. GraphGPT~\cite{tang2024graphgptgraphinstructiontuning} aligns LLMs with graph structures to improve document understanding. GraphAdapter~\cite{huang2024gnngoodadapterllms} uses LLMs on graph-structured data with parameter-efficient tuning, yielding significant improvements in node classification. LLM-GNN~\cite{chen2023label} leverages LLMs to annotate node labels, which are subsequently used to train Graph Convolutional Networks (GCNs) in an instruction-tuning paradigm. OFA~\cite{liu2024alltraininggraphmodel} introduces a novel graph prompting paradigm that appends prompting substructures to input graphs, enabling it to address various tasks, including node classification, without the need for fine-tuning. ZeroG~\cite{li2024zeroginvestigatingcrossdatasetzeroshot} uses language models to encode both node attributes and class semantics, achieving significant performance improvements on node classification tasks.

These advancements in text-attributed graph methods have been successfully applied to a range of tasks, including text classification~\cite{wen2023augmenting,zhang2021semi}, citation recommendation~\cite{bai2018neural, xie2021graph}, question answering~\cite{yasunaga2022linkbert}, and document retrieval~\cite{ma2021pre}.

\subsection{Attributed Graph Clustering}
Attributed graph clustering (AGC) aims to effectively leverage both structural and attribute information in graphs for improved clustering performance~\cite{yang2021effective}, which has been extensively studied in the literature~\cite{bothorel2015clustering,liu2022survey,chunaev2020community,li2024versatile,yang2021effective,li2023efficient,lin2024spectral,yang2024effective}.
DAEGC~\cite{wang2019attributed} uses attention mechanisms to adaptively aggregate neighborhood information, improving the expressiveness of node embeddings. AGCN~\cite{peng2021attention} dynamically blends attribute features from autoencoders (AE) with topological features from GCNs, using a heterogeneous fusion module to integrate both types of information.
DFCN ~\cite{tu2021deep} combines representations from AE and GAE hidden layers through a fusion module and employs a triple self-supervision strategy to enhance cross-modal information utilization. CCGC ~\cite{yang2023cluster} employs contrastive learning with non-shared weight Siamese encoders to construct multiple views, improving the robustness and reliability of clustering through high-confidence semantic pairings. AGC-DRR ~\cite{gong2022attributed}, reduces redundant information in both input and latent spaces, enhancing the network's robustness and feature discriminability.

\subsection{Difference from Previous Works}

In this section, we outline the key differences between our method and prior label-free approaches~\cite{chen2023label,li2024enhancinggraphneuralnetworks}. Specifically, \cite{chen2023label} employs traditional active selection techniques, such as FeatProp, RIM, and GraphPart, for node selection, performing annotation only once before the GNN training phase. In contrast, \algo introduces a novel subspace clustering approach and conducts annotation in batches throughout the self-training process. Additionally, \algo leverages reliable GNN predictions as pseudo-labels, thereby reducing the need for expensive LLM queries, and incorporates a Hybrid Label Refinement module to enhance the quality of node selection.

Ref.~\cite{li2024enhancinggraphneuralnetworks}, while employing iterative node selection for annotation, initializes the process with random node selection, unlike our method, which utilizes subspace clustering. Furthermore, the selection metric defined in \cite{li2024enhancinggraphneuralnetworks} differs significantly from ours. Lastly, their approach relies on the LLM to explain annotation decisions for knowledge distillation, which substantially increases the query cost, whereas \algo mitigates this by merely generating the annotation result with confidence.

\section{Preliminaries}

\subsection{Problem Statement}\label{sec:problem}
Let $\G=(\V,\EDG, \T)$ be a {\em text-attributed graph}, wherein $\V$ stands for a set of $n$ nodes and $\EDG$ represents a set of $m$ edges between nodes in $\V$. For each edge $(v_i,v_j)\in \EDG$, we say $v_i$ and $v_j$ are neighbors to each other and use $\N(v_i)$ to denote the set of neighbors of $v_i$. Each node $v_i$ in $\G$ is characterized by a text description $T_i$ in $\T$. We denote by $\AM$ the adjacency matrix of $\G$, in which $\AM_{i,j}=\AM_{j,i}=1$ if $(v_i,v_j)\in \EDG$ and $0$ otherwise. 
Accordingly, $\LM=\DM-\AM$ is the Laplacian matrix of $\G$, where $\DM$ is the diagonal degree matrix satisfying $\DM_{i,i}=|\N(v_i)|\ \forall{v_i\in \V}$. $\NAM=\DM^{-1/2}\AM\DM^{-1/2}$ stands for the normalized version of $\AM$ and $\tilde{\LM}=\IM-\NAM$ is used to symbolize the normalized Laplacian of $\G$.

Let $\C=\{c_1,c_2,\cdots,c_k\}$ be a set of $k$ classes, where each class is associated with a label text.
Given a TAG $\G=(\V,\EDG, \T)$ and $k$ classes $\C$, the goal of {\em label-free node classification}~\cite{chen2023label,li2023prompt} is to predict the class labels of {\em all} nodes in $\V$. 

\subsection{Graph Neural Networks}
The majority of existing GNNs~\cite{kipf2016semi,gasteiger2018predict,xu2018representation,chen2020simple,wu2019simplifying} mainly follow the {\em message passing} paradigm~\cite{gilmer2017neural}, which first aggregates features form the neighborhood, followed by a transformation. 
As demystified in recent studies~\cite{Ma2020AUV,Zhu2021InterpretingAU}, after removing non-linear operations, graph convolutional layers in popular {\em graph neural network} models, e.g., APPNP~\cite{gasteiger2018predict}, GCNII~\cite{chen2020simple}, and JKNet~\cite{xu2018representation}, essentially optimize the {\em graph Laplacian smoothing}~\cite{dong2016learning} problem as formulated in Eq.~\eqref{eq:GLS-mat}.
\begin{equation}\label{eq:GLS-mat}
\min_{\HM}{(1-\alpha)\cdot\|\HM - \XM\|^2_F} + \alpha \cdot \Tr(\HM^\top \tilde{\LM} \HM),
\end{equation}
where $\alpha\in [0,1]$ is a coefficient balancing two terms. The first term in Eq.~\eqref{eq:GLS-mat} seeks to reduce the discrepancy between the input matrix $\XM$ and the target node representations $\HM$. 
The second term calculates the {\em Dirichlet Energy}~\cite{chung1997spectral} of $\HM$ over $\G$, which can be rewritten as $\footnotesize \sum_{(v_i,v_j)\in \EDG}{\left\|\frac{\HM_{i}}{\sqrt{d(v_i)}}-\frac{\HM_{j}}{\sqrt{d(v_j)}}\right\|^2_2}$, enforcing $\HM_i$, $\HM_j$ of any adjacent nodes $(v_i,v_j)\in \EDG$ to be close. 
By taking the derivative of Eq.~\eqref{eq:GLS-mat} w.r.t. $\ZM$ to zero and applying Neumann series~\cite{horn2012matrix}, the closed-form solution (i.e., final node representations) can be expressed as
\begin{equation}\label{eq:smooth-x}
\HM = \sum_{t=0}^{\infty}{(1-\alpha)\alpha^t\NAM^t}\XM.
\end{equation}

\subsection{Large Language Models and Prompting}
In this work, we refer to LLMs as the language models that have been pre-trained on extensive text corpora, which exhibit superb comprehension ability and massive knowledge at the cost of billions of parameters, such as LLaMA~\cite{touvron2023llama} and GPT4~\cite{achiam2023gpt}.
The advent of LLMs has brought a new paradigm for task adaptation, which is known as ``pre-train, prompt, and predict''. In such a paradigm, instead of undergoing cumbersome model fine-tuning on task-specific labeled data, the LLMs pre-trained on a large text corpus are queried with a natural language prompt specifying the task and context, and the models return the answer based on the instruction and the input text~\cite{liu2023pre}. 
For example, given a paper titled ``BERT: Pre-training of Deep Bidirectional Transformers for Language Understanding'' and the task of predicting its subject. The prompts for this task can be:
\begin{equation*}
\text{\{[Title], this, paper, belong, to, which, subject?\}}
\end{equation*}
While querying a model once with a reasonably sized input is affordable for most users, challenges arise when the input text is exceptionally long, such as a complete book or a lengthy article from a website. Additionally, the need to query large datasets further escalates costs. For instance, the \textsc{PubMed} dataset contains tens of thousands of articles, with the longest exceeding 100,00 words. A full-scale prediction of all article subjects in the \textsc{PubMed} dataset using GPT-3.5 could cost over 20 dollars, and with the more advanced GPT-4, this cost rises to around 400 dollars\footnote{Price estimates are based on gpt-3.5-turbo-16k-0613 and gpt-4-32k-0613 models.}.

Apart from incurring high costs, LLM predictions can sometimes be noisy. Although these models may exhibit high confidence in their outputs, they are prone to errors, a phenomenon often referred to as AI hallucination~\cite{rawte2023survey}. One potential solution to mitigate this issue is to experiment with different prompts and select the most accurate output. However, each attempt incurs significant costs due to the expensive nature of these models.

\section{Methodology}

This section presents our \algo framework that integrates LLMs into the GNNs adaptively for label-free node classification. 

\begin{figure*}
\centering
\includegraphics[width=0.9\textwidth]{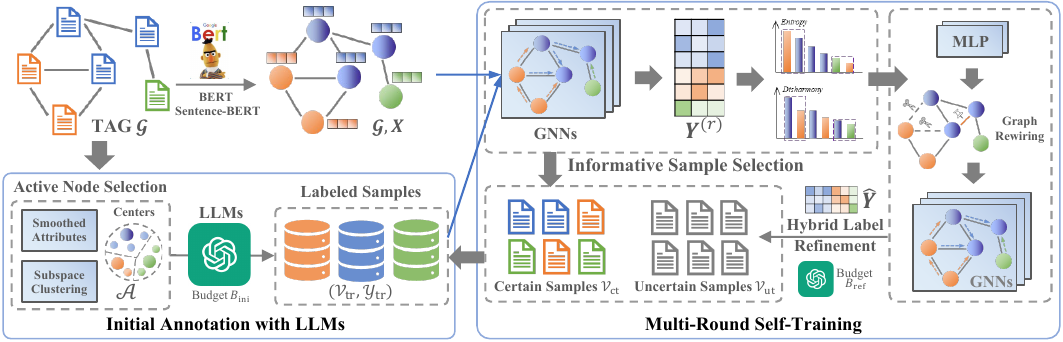}
\vspace{-2ex}
\caption{Pipeline of Our Proposed \algo}
\label{fig:pipeline}
\end{figure*}

\subsection{Synoptic Overview of \algo}\label{sec:overview}

The pipeline of \algo is illustrated in Figure~\ref{fig:pipeline}, which mainly works in two stages: initial node annotations with LLMs (Stage I) and multi-round self-training based on GNNs (Stage II). 
More specifically, given a TAG $\G$, \algo first converts $\G$ into a standard attributed graph by encoding the text description $T_i$ of each node $v_i\in \V$ into a text embedding $\XM_i\in \mathbb{R}^d$ as its attribute vector using PLMs, e.g., BERT~\cite{devlin2019bertpretrainingdeepbidirectional} and Sentence-BERT~\cite{reimers2019sentencebertsentenceembeddingsusing}. 

Subsequently, \algo proceeds to the first stage, which seeks to create pseudo-labels $\YTR$ for a small set $\VTR$ of nodes using LLMs as initial training samples. Given an allocated budget of $\BPA=\varepsilon\cdot B$ ($0< \varepsilon\le 1$) for querying LLMs, where $B$ is the total query budget, Stage I extracts a small set $\A$ of $\BPA$ representative nodes from $\V$ by our {\em active node selection strategy}, followed by carefully-designed prompts to LLMs for accurate annotations of samples in $\A$.

After that, \algo starts Stage II, which trains GNNs for $R$ rounds by taking as input the graph $\G$ with attribute matrix $\XM$, training samples $(\VTR,\YTR)$, and the remaining query budget of $\BLC = (1-\varepsilon)\cdot B$ to LLMs.
In $r$-th ($1\le r\le R$) round of the self-training, \algo begins by generating current label predictions $\YM^{(r)}\in \mathbb{R}^{|\V|\times k}$ by a GNN model trained on $(\VTR,\YTR)$. With the class probabilities therefrom and the graph structures, our {\em informative sample selection} module first evaluates the {\em informativeness} of nodes in the test set $\V\setminus \VTR$ and then filter out two small sets $\VCS$ and $\VUS$ that comprise samples for which the current model is the most and least confident and about their class predictions, respectively.
For the most uncertain samples in $\VUS$, we resort to a {\em hybrid} approach for refining or rectifying their predicted labels, which effectively combines additional knowledge injected from LLMs and rewired graph topology.
In the end, \algo expands $\VTR$ with $\VCS$ and $\VUS$ and enters into next round of self-training. 

In what follows, we first introduce our strategies for selecting active nodes and querying LLMs for initial node annotations in Section~\ref{sec:initial}. Sections~\ref{sec:sample-selection} and ~\ref{sec:hybrid} elaborate on the selection of informative samples and our hybrid label refinement scheme, respectively.
{In Sections~\ref{sec:training} and ~\ref{sec:analysis}, we describe the training objectives of \algo and conduct related analyses.}



\subsection{Initial Node Annotations}\label{sec:initial}
This stage involves the selection of nodes $\A$ for annotation and specific labeling tricks using LLMs.

\subsubsection{\bf Active Node Selection}
\label{sec:selection}
The basic idea of our node selection strategy is to partition nodes in $\V$ into a number of clusters and select a small set $\A$ ($|\A|=\BPA$) of distinct cluster centers and nodes as the representatives for subsequent annotations, referred to as the {\em active node set} (ANS).

To extract ANS $\A$, we first calculate a $T$-truncated approximation of the GNN-based node representations in Eq.~\eqref{eq:smooth-x} by
\begin{equation}\label{eq:initial_combine}
\HM = \sum_{t=0}^{T}{(1-\alpha)\alpha^t \NAM^t \XM}
\end{equation}
as the feature vectors of nodes, which encode both textual and structural semantics underlying $\G$.
Next, \algo opts for the {\em subspace clustering} (SC) technique~\cite{vidal2011subspace,lin2024spectral} for node grouping, which is powerful in noise reduction and discovering structural patterns underlying the feature vectors~\cite{fettal2023scalable}.
To be specific, SC first constructs an affinity graph (a.k.a. self-expressive matrix), i.e., $\SM$, from feature vectors $\HM$ such that the following objective is optimized:
\begin{equation}\label{eq:obj-sc}
\min_{\SM\in \mathbb{R}^{n\times n}}  \|\HM - \SM \HM\|_F^2 + \Omega(\SM),
\end{equation}
where $\Omega(\SM)$ signifies a regularization term introduced to impose additional structure constraints on $\SM$. A popular choice for $\Omega(\SM)$ is the nuclear norm $\|\SM\|_{*}$, which is to promote the low-rankness of $\SM$~\cite{liu2010robust}.
As such, if we let $\UM$ be the left singular vectors of $\XM$, the minimizer of Eq.~\eqref{eq:obj-sc} is uniquely given by $\SM=\UM\UM^{\top}$~\cite{liu2010robust}. 
\begin{lemma}\label{lem:sc-kmeans}
The spectral clustering of $\SM$ with $K$ desired clusters is equivalent to applying the $K$-Means over $\UM$.
\end{lemma}
Afterwards, standard SC applies the {\em spectral clustering}~\cite{von2007tutorial} over $\frac{\SM+\SM^{\top}}{2}$ to derive clusters. Lemma~\ref{lem:sc-kmeans}\footnote{All missing proofs appear in the Appendix of our extended version.} establishes a connection between the spectral clustering of $\SM$ and $K$-Means over $\UM$.
Note that the exact construction of $\UM$ entails prohibitive computational and space costs of $O(|\V|^2)$.
As a partial remedy, we simplify the clustering as executing the $K$-Means with top-$\tau$ (typically $\tau=128$ ) left singular vectors $\UM^{(\tau)}$ of $\XM$. 

Denote by $u_1,\ldots,u_K\in \V$ the $K$ centers obtained. 
We sort the remaining nodes in $\V\setminus \{u_1,\ldots,u_K\}$ in descending order by their closeness to their respective centers as defined in Eq.~\eqref{eq:NCC}, and add the top $\BPA-K$ ones together with $u_1,\ldots,u_K$ to ANS $\A$.
\begin{equation}\label{eq:NCC}
\frac{1}{1 + \left\|\UM^{(\tau)}_{i} - \UM^{(\tau)}_{u_j}\right\|_2}
\end{equation}
The rationale is based on the observation from~\cite{chen2023label} indicating that nodes closer to cluster centers show higher quality and lower difficulty in annotation.

\subsubsection{\bf LLM-based Annotation}\label{sec:LLM-annotate}
Given the ANS $\A$, \algo then query LLMs (e.g., GPT-3.5) for generating the annotation and confidence score for each node in $\A$ using the {\em consistency prompt} strategy~\cite{wangself} adopted in \cite{chen2023label}. 
Given these confidence scores, a post-filtering~\cite{chen2023label} is further applied to filter out the low-confidence samples with low-quality labels.
Finally, we utilize the ANS $\A$ and their annotations as the initial training data ($\VTR,\YTR$) input to the \algo model for self-training.
We defer detailed prompt descriptions and examples to the Appendix of our extended version.






\subsection{\bf Informative Sample Selection}\label{sec:sample-selection}
As delineated in Section~\ref{sec:overview}, in the course of self-training, \algo selects node samples from the test node set $\V\setminus \VTR$ with high informativeness for the subsequent training based on the feedback provided by the past and current models trained with $(\VTR, \YTR)$. 

Given the predictive probabilities in $\YM^{(r)}$ output by current model $f^{(r)}$, $\V\setminus \VTR$ can be roughly categorized into three groups of node samples: {\em most certain samples} $\VCS$, {\em most uncertain samples} $\VUS$, and others.
To be precise, $\VCS$ consists of nodes for which we are most certain about their label predictions in $\YM^{(r)}$, while $\VUS$ includes the most uncertain ones.
Intuitively, node samples in $\VCS$ and $\VUS$ are the most informative ones. The former can be used as additional supervision signals that guide the model towards accurate predictions, while the latter causes the performance {\em bottleneck} of the current model $f^{(r)}$, whose predictions are tentative due to data scarcity or limited model capacity.
Accordingly, identifying such nodes for label refinement is crucial for upgrading the model.
Next, we attend to constructing $\VCS$ and $\VUS$. In particular, we set $|\VCS|$ as a hyper-parameter, and $|\VUS| = \frac{\BLC}{R-1}$.



Firstly, we employ a GNN as the backbone $f^{(r)}$, followed by a softmax output layer, to obtain the new classification result:
\begin{equation}\label{eq:Yr}
\YM^{(r)} = \textsf{softmax}(\textsf{GNN}(\G, \XM)).
\end{equation}
To alleviate the model bias and sensitivity in each round and ensure a robust and accurate identification of $\VCS$ and $\VUS$, we propose an ensemble scheme that averages the label predictions in past rounds to reach a consensus as follows:
\begin{equation}
\overline{\YM}^{(r)} = \sum_{\ell=1}^{r}{\frac{(1-\alpha)\alpha^\ell}{1-\alpha^r}\YM^{(\ell)}},
\end{equation}
where $\frac{(1-\alpha)\alpha^\ell}{1-\alpha^r}$ stands for the weight assigned for the $\ell$-th round result, which decreases as $\ell$ increases since intuitively initial predictions are less accurate than the latter ones.
On its basis, we introduce two metrics, {\em label entropy} and {\em label disharmonicity}, to quantify the certainty or uncertainty of node samples.

\stitle{Label Entropy} A simple and straightforward way is to adopt the prominent Shannon {\em entropy}~\cite{shannon1948mathematical} from the information theory to measure the uncertainty. Mathematically, the {\em label entropy} of a node $v_i\in \V\setminus \VTR$ can be formulated as
\begin{equation}\label{def:entropy}
LE(v_i) = -\sum_{j=1}^{k} \overline{\YM}^{(r)}_{i,j} \cdot \log \left( \overline{\YM}^{(r)}_{i,j} + \sigma \right),
\end{equation}
where $\sigma$ (typically $10^{-9}$) is introduced to avoid zero values. 
A high label entropy value connotes that the predicted probabilities are evenly distributed among all possible classes, and hence, indicates that the prediction is highly uncertain. Intuitively, the larger $LE(v_i)$ is, the less confident we are about the label predictions for node $v_i$.

\stitle{Label Disharmonicity} 
Note that the label entropy assesses the predictive outcome by merely inspecting the node-class relations in $\YM^{(r)}$, which disregards the correlations between samples, i.e., topological connections in $\G$, and thus, engenders biased evaluation.
To remedy its deficiency, we additionally introduce a novel notion of {\em label disharmonicity}. For each node $v_i\in \V\setminus \VTR$, its label disharmonicity is defined by
\begin{small}
\begin{equation}\label{def:disharmonicity}
LH(v_i) = \sqrt{\sum_{j=1}^{k}{\left(\overline{\YM}^{(r)}_{i,j}-\frac{1}{|\N(v_i)|}\sum_{v_{\ell}\in \N(v_i)}{\overline{\YM}^{(r)}_{\ell,j}}\right)^2}},
\end{equation}
\end{small}
which calculates the 
the discrepancy of $v_i$'s class labels from those of its neighbors in $\G$. 
Intuitively, the larger $LH(v_i)$ is, the more unreliable the predicted label of $v_i$ by $\overline{\YM}^{(r)}$ is.

\stitle{Constructions of $\VCS$ and $\VUS$}
With the foregoing scores at hand, we create $\VCS$ as follows. Firstly, we extract two sets $\Sset_{\text{har}}$ and $\Sset_{\text{ent}}$ from $\V\setminus \VTR$ by
\begin{equation}\label{eq:SS-topk}
\Sset_{\text{har}} = \argtopk{v_i\in \V\setminus \VTR}{-LH(v_i)}\ \text{and}\ \Sset_{\text{ent}} = \argtopk{v_i\in \V\setminus \VTR}{LE(v_i)},
\end{equation}
where $k=\frac{|\BLC|}{R-1}$ and $\Sset_{\text{har}}$ (resp. $\Sset_{\text{ent}}$) contains the nodes with $\frac{|\BLC|}{R-1}$-largest (resp. smallest) label entropy (resp. disharmonicity) in $\V\setminus \VTR$.
In turn, the set $\VCS$ can be formed via
\begin{equation*}
\VCS = \left( \Sset_{\text{har}} \cap \Sset_{\text{ent}} \right) \cup \overline{\Sset},
\end{equation*}
where $\overline{\Sset}$ is a union of the remaining samples picked from $\Sset_{\text{har}}$ and $\Sset_{\text{ent}}$ with $k=\left({|\BLC|}/{(R-1)}-|\left( \Sset_{\text{har}} \cap \Sset_{\text{ent}} \right)|\right)/2$ as follows:
\begin{equation}\label{eq:SS-topk-Union}
\overline{\Sset} = \left(\argtopk{v_i\in \Sset_{\text{har}}\setminus \Sset_{\text{ent}}}{-LH(v_i)}\right) \cup \left(\argtopk{v_i\in \Sset_{\text{ent}}\setminus \Sset_{\text{har}}}{LE(v_i)}\right).
\end{equation}
In a similar vein, we can obtain $\VUS$ by sorting the nodes in reverse orders, i.e., replacing $-LH(v_i)$ and $LE(v_i)$ in Eq.~\eqref{eq:SS-topk} and Eq.~\eqref{eq:SS-topk-Union} by $LH(v_i)$ and $-LE(v_i)$, respectively.





\subsection{Hybrid Label Refinement}\label{sec:hybrid}

As remarked in the preceding section, $\VUS$ consists of the bottleneck samples where the past and current GNN models largely fail. 
The uncertain predictions for samples in $\VUS$ can be ascribed to two primary causes. First, there is a lack of sufficient representative labeled samples for model training. Second, the nodes in $\VUS$ are connected to scarce or noisy edges that can easily mislead the GNN inference.
Naturally, it is necessary to seek auxiliary label information from LLMs.
However, as revealed in~\cite{chen2023label}, the predictions made by LLMs can also be noisy, particularly for instances in $\VUS$ with low confidence. 
To mitigate this issue, we propose to rewire the graph structure surrounding the uncertain samples and re-train a GNN model to get their new predictions $\widehat{\YM}^{(r)}$. On top of that, we refine the labels of node samples in $\VUS$ through a careful combination of the signals from LLMs and $\widehat{\YM}^{(r)}$.


\subsubsection{\bf Graph Rewiring-based Predictions}\label{sec:graph-rewire}
To eliminate noisy topology and complete missing links, we propose to align the graph structure with the node features. Recall that in Eq.~\eqref{eq:GLS-mat}, GNNs aim to make node features $\HM$ optimize the Dirichlet Energy of node features over normalized graph Laplacian $\tilde{\LM}$, thereby enforcing the feature vectors of adjacent nodes to be similar. Conversely, our idea is to optimize the Dirichlet Energy by updating $\tilde{\LM}$ while fixing $\HM$.
\begin{lemma}\label{lem:laplacian}
$\textstyle \IM-\frac{{\beta}\cdot\HM\HM^{\top}}{\|\HM\HM^\top\|_F} = \underset{{\tilde{\LM}}}{\arg\min}{\Tr(\HM^{\top}\tilde{\LM}\HM)}$ s.t. $\|\NAM\|_F=\beta$.
\end{lemma}
Along this line, we can obtain a new graph $\HH$ with an adjacency matrix $\HM\HM^{\top}$ by Lemma~\ref{lem:laplacian}. The connections therein can be used to complement the input graph topology in $\G$.

More concretely, \algo first generates the node feature through an MLP layer as follows:
\begin{equation*}
\HM = \textsf{MLP}(\G, \YM^{(r)}).
\end{equation*}
The graph $\HH$ thus can be constructed by assigning each edge $(v_i,v_j)$ a non-negative weight $w(v_i,v_j)$ computed by
\begin{equation*}
w(v_i,v_j) = \max(\HM_{i}\cdot\HM_{j}^{\top},0).
\end{equation*}
As in Eq.~\eqref{eq:add-rm-edges}, \algo forms (i) a set $\EDG^{(-)}$ of edges to be removed from $\G$, and (ii) a set $\EDG^{(+)}$ of node pairs to be connected in $\G$, by picking edges from $\EDG$ with the $\delta^{(-)}\cdot |\EDG|$-smallest weights and node pairs from $\VTR\times \V\setminus \VTR$ with the $\delta^{(+)}\cdot |\EDG|$-largest weights, respectively, where $\delta^{(-)}$ and $\delta^{(+)}$ are ratio parameters.
\begin{small}
\begin{equation}\label{eq:add-rm-edges}
\EDG^{(-)} = \argtopk{(v_i,v_j)\in \EDG}{-w(v_i,v_j)},\ \EDG^{(+)} = \argtopk{\substack{v_i\in \VTR, v_j\in \V\setminus\VTR\\ (v_i,v_i)\notin \EDG}}{w(v_i,v_j)}.
\end{equation}
\end{small}
The intuition for the way of creating $\EDG^{(+)}$ is that by connecting unlabeled nodes to their similar samples with certain labels, the GNN model is empowered to infer their labels more accurately and confidently.


Then, we construct a rewired graph $\widehat{\G}$ with edge set $(\EDG\cup \EDG^{(+)})\setminus \EDG^{(-)}$ and derive an updated label predictions in Eq.~\eqref{eq:new-Y} based thereon.
\begin{equation}\label{eq:new-Y}
\widehat{\YM}^{(r)} = \textsf{softmax}(\textsf{GNN}(\widehat{\G}, \YM^{(r)}))
\end{equation}


\subsubsection{\bf Label Refinement with LLMs and $\widehat{\YM}^{(r)}$}
For each node $v_i$ in $\VUS$, \algo requests its annotation $y_{i}$ with a confidence score $\phi(v_i)$ from LLMs as in Section~\ref{sec:LLM-annotate}. Similarly, by $\widehat{\YM}^{(r)}$ obtained in Eq.~\eqref{eq:new-Y}, we can get the most possible label $\hat{y}_i$ for node $v_i\in \VUS$.
If $y_i\neq \hat{y}_i$, we should keep the one that we are more confident about. More precisely, let $rank_{\text{LLM}}(v_i)$ and $rank_{\text{GNN}}(v_i)$ be the rank of node $v_i$ within $\VUS$ according to their confidence scores $\phi(v_i)$ or prediction probabilities in $\widehat{\YM}^{(r)}$, respectively. 
Let $\overline{\phi}$ be a predefined threshold, indicating the minimum confidence score to trust the result by LLMs.
If $rank_{\text{LLM}}(v_i) < rank_{\text{GNN}}(v_i)$ or $\phi(v_i)\le \overline{\phi}$, \algo is less certain about the annotation by LLMs, and hence, we set $\hat{y}_i$ as the final label for $v_i$.




\subsection{Model Optimization}\label{sec:training}
In each round of self-training, we train the \algo model by optimizing two objectives pertinent to classification and rewired topology with the current training samples $\VTR$ and their labels $\YTR$.
Following common practice, we adopt the {\em cross-entropy} loss for node classification, which is formulated by
\begin{equation*}
\mathcal{L}_{\text{CLS}} = \textsf{CrossEntropy}(\YM_{\text{tr}}, \widehat{\YM}^{(r)}_{\text{tr}}),
\end{equation*}
where $\YM_{\text{tr}}$ stands for the ground-truth labels for nodes in $\VTR$ and $\widehat{\YM}^{(r)}_{\text{tr}}$ contains the class probabilities of training samples predicted over the rewired graph $\widehat{\G}$ as in Eq.~\eqref{eq:new-Y}.

To align with our objective of graph rewiring in Section~\eqref{sec:graph-rewire}, we additionally include 
the following loss:
\begin{equation}
   \mathcal{L}_{\text{DE}} = \frac{|V|}{|E|}\cdot\left(\Tr\left({\YM^{(r)}}^\top \tilde{\LM}_{\widehat{\G}} \YM^{(r)}\right) - \lambda\cdot \| \tilde{\LM}_{\widehat{\G}} \|^2_F \right),
   \label{eq:enc-loss}
\end{equation}
where the first term measures the Dirichlet Energy of label predictions over the rewired graph $\widehat{\G}$, while the latter is the Tikhonov regularization~\cite{tikhonov1977solutions} introduced to avoid trivial solutions.
$\lambda$ denotes the trade-off parameter between two terms. 

\subsection{Theoretical Analyses}\label{sec:analysis}

\subsubsection{\bf Connection between Label Disharmonicity and Dirichlet Energy} Given a graph signal $\boldsymbol{x}\in \mathbb{R}^n$, its standard Dirichlet Energy on $\G$ is $\boldsymbol{x}^{\top}\LM\boldsymbol{x}$, whose gradient can be expressed as $\LM\boldsymbol{x}$. Accordingly, for the probabilities of all nodes w.r.t. $j$-th class in $\overline{\YM}^{(r)}_{\cdot,j}$, the gradient of its Dirichlet Energy is $\LM\overline{\YM}^{(r)}_{\cdot,j}$, wherein each $i$-th entry equals $\textstyle \sum_{v_\ell\in \N(v_i)}{\left(\overline{\YM}^{(r)}_{i,j}-\overline{\YM}^{(r)}_{\ell,}\right)}$.
As per the definition of the label disharmonicity in Eq.~\eqref{def:disharmonicity}, it is easy to prove that $LH(v_i)=\frac{1}{\sqrt{|\N(v_i)|}}\cdot\|(\LM\overline{\YM}^{(r)})_i\|_2$. Namely, the label disharmonicity of a node $v_i$ is the reweighted $L_2$ norm of its Dirichlet Energy gradients of all classes over $\G$.

\subsubsection{\bf Connection between $\mathcal{L}_{\text{DE}}$ and Spectral Clustering}
{Spectral clustering}~\cite{von2007tutorial} seeks to partition nodes in graph $\G$ into $K$ disjoint clusters $\{\C_1,\ldots,\C_K\}$ such that their intra-cluster connectivity is minimized. A common formulation of such an objective is the RatioCut~\cite{hagen1992new}: $\min_{\{\C_1,\ldots,\C_K\}}{\sum_{k=1}^{K}{\frac{1}{K}\sum_{v_i\in \C_k, v_j\in \V\setminus \C_k}{\frac{\NAM_{i,j}}{|\C_k|}}}}$, which is equivalent to
\begin{equation}\label{eq:spectral}
\min_{\CM}\Tr(\CM^{\top}(\IM-\NAM)\CM)=\Tr(\CM^{\top}\tilde{\LM}\CM).
\end{equation}
$\CM \in \mathbb{R}^{n\times K}$ denotes a node-cluster indicator matrix in which $\CM_{i,j}=\frac{1}{\sqrt{|\mathcal{C}_j|}}$ if $v_i\in \mathcal{C}_j$, and $0$ otherwise. This discreteness condition on $\CM$ is usually relaxed in standard spectral clustering and $\CM$ is allowed to be a continuous probability distribution. 
Let $\CM={\YM^{(r)}}$ and $\tilde{\LM}=\tilde{\LM}_{\widehat{\G}}$. Our Dirichlet Energy term in Eq.~\eqref{eq:enc-loss} is a RatioCut in spectral clustering in essence.



\section{Experiments}\label{sec:exp}
In this section, we experimentally evaluate \algo in label-free node classification over five real TAGs. 
Particularly, we investigate the following research questions: \textbf{RQ1:} How is the performance of \algo compared to the existing zero-shot/label-free methods? 
\textbf{RQ2:} How do the three proposed components of \algo affect its performance? 
\textbf{RQ3:} How do the key parameters in \algo affect its performance? 
\textbf{RQ4:} What are the computational and financial costs of \algo?

\begin{table}[!t]
\centering
\caption{Dataset statistics.}
\label{tab:stats_brief}
\vspace{-2ex}
\begin{small}
\begin{tabular}{c|c|c|c|c}
\hline
{\bf Dataset} & {\bf \#Nodes} & {\bf \#Edges} & {\bf \#Classes} & {\bf Type} \\ \hline
{\em Cora}~\cite{mccallum2000automating} & $2,708$ & $5,429$ & 7 & Citation Graph \\ 
{\em Citeseer}~\cite{giles1998citeseer} & $3,186$ & $4,277$ & 6 & Citation Graph \\ 
{\em Pubmed}~\cite{sen2008collective} & $19,717$ & $44,335$ & 3 & Citation Graph \\ 
{\em WikiCS}~\cite{mernyei2020wiki} & $11,701$ & $215,863$ & 10 & Hyperlink Graph \\ 
{\em DBLP}~\cite{ji2010graph} & $14,376$ & $431,326$ & 4 & Citation Graph \\ \hline
\end{tabular}
\end{small}
\label{tab:datasets}
\vspace{-1ex}
\end{table}

\subsection{Experiment Settings}

\begin{table*}[!t]
\centering
\renewcommand{\arraystretch}{1.0}
\caption{Label-free node classification performance.}\vspace{-3mm}
\begin{small}
\addtolength{\tabcolsep}{-0.25em}
\resizebox{\textwidth}{!}{%
\begin{tabular}{c|c|c c c c|c c c c|c c c c|c c c c|c c c c}
\hline
\multirow{2}{*}{\bf Backbone} & \multirow{2}{*}{\bf Method} & \multicolumn{4}{c|}{\bf{ {\em Cora}}} & \multicolumn{4}{c|}{\bf{ {\em CiteSeer}}} & \multicolumn{4}{c|}{\bf{ {\em PubMed}}} & \multicolumn{4}{c|}{\bf{ {\em WikiCS}}}  & \multicolumn{4}{c}{\bf{ {\em DBLP}}}  \\ \cline{3-22}
& & Acc & NMI & ARI & F1 & Acc & NMI & ARI & F1 & Acc & NMI & ARI & F1 & Acc & NMI & ARI & F1  & Acc & NMI & ARI & F1 \\ \hline
\multirow{2}{1cm}{\centering \texttt{Bert}~\cite{devlin2019bertpretrainingdeepbidirectional}} &  Base &29.36&0.38&-0.19&7.86&19.27&0.44&0.28&11.58&20.83&0&0&19.15&7.71&1.61&0.43&9.88&18.41&0.20&-0.07&14.08 \\ 
& \texttt{Sentence-BERT}~\cite{reimers2019sentencebertsentenceembeddingsusing} &56.20&35.68&32.84&54.06&45.35&30.19&26.1&44.51&32.71&14.08&1.02&42.51&59.95&37.52&32.75&57.16&55.84&23.11&16.89&55.75\\  \hline
\multirow{1}{1cm}{\centering \texttt{Bart}~\cite{lewis2019bartdenoisingsequencetosequencepretraining}} & 
 \texttt{BART-Large-MNLI} &40.88&16.62&12.44&39.07&42.81&16.7&14.66&38.29&67.60&29.37&29.42&66.64&46.69&24.51&19.61&43.28&49.06&8.79&9.46&44.67\\ 
 \hline
 \multirow{2}{2cm}{\centering \texttt{LLM as Predictor}~\cite{chen2024exploring}} 
 & Zero-shot  
     &64.00&45.47&38.03&- &63.00&42.66&35.91&-&\textbf{86.00}&\textbf{59.51}&\textbf{65.78}&-&66.00&\textbf{57.21}&47.42&-&64.00&30.10&31.56&- \\
     & Zero-shot (COT)
     &68.00&46.38&40.22&-&65.50&46.18&37.35&-&83.00&\underline{56.02}&\underline{61.21}&-&65.50&53.65&47.72&-&67.50&31.61&34.42&- \\
     
     \hline
       \multirow{3}{2cm}{\centering \texttt{Clustering}} 
&{AGC-DRR}~\cite{gong2022attributed} &66.97&53.66&46.14&63.21&69.56&48.54&49.42&65.97&65.87&25.10&25.78&65.48&57.17&46.06&39.89&48.46&74.04&44.17&45.98&72.66 \\
& {CCGC}~\cite{yang2023cluster}&\underline{74.62}&\underline{56.71}&\underline{53.86}&\underline{71.46}&49.45&33.23&35.11&44.29&44.16&17.74&17.68&44.14&32.08&21.06&7.66&30.76&72.85&40.92&45.02&70.61 \\
& {DAEGC}~\cite{wang2019attributed}&72.62&53.40&49.51&70.06&69.08&\underline{47.04}&\underline{47.55}&65.21&62.97&22.61&21.15&63.59&59.09&47.30&45.72&52.64&\underline{75.04}&\underline{43.70}&\underline{48.92}&\underline{72.74} \\

     \hline
      \multirow{2}{2cm}{\centering \texttt{PLM+GNN}} 
 & OFA~\cite{liu2024alltraininggraphmodel}
     &29.00&6.92&4.98&31.55 &41.80&12.72&7.97&39.45&29.80&0.41&0.18&40.49&25.80&10.11&4.87&28.60&44.00&4.56&3.08&42.21\\
     & ZeroG~\cite{li2024zeroginvestigatingcrossdatasetzeroshot}
    &61.12&44.79&44.11&56.16 &46.53&27.33&25.74&41.84&75.82&35.57&42.48&73.62&47.92&37.01&26.72&39.48&58.10&24.73&27.84&54.70 \\
     \hline
\multirow{5}{1cm}{\centering \texttt{GCN}~\cite{kipf2016semi}}  & \texttt{LLM-GNN} (FP) & 72.52 & 54.83 & 52.56 & 68.77 &66.81&44.46&40.23&63.88&79.90&42.44&48.79&77.97&65.75&51.06&47.34&54.43&70.19&38.06&38.48&69.00 \\ 
& \texttt{LLM-GNN} (GP) &71.20&53.88&48.67&67.76&\underline{70.86}&46.20&46.00&\underline{66.04}&81.97&45.70&52.97&80.92&66.26&50.51&\underline{48.57}&58.21&70.53&37.29&39.46&68.90 \\  & \texttt{LLM-GNN} (RIM) &72.34&52.46&52.47&68.55&66.51&42.15&39.09&62.63&82.41&46.62&55.08&81.21&66.44&49.50&46.87&54.74&73.15&42.16&43.69&71.81 \\ 
\cline{2-22}
& \algo &\cellcolor{gray!30} \textbf{81.80}&\cellcolor{gray!30} \textbf{64.22}&\cellcolor{gray!30} \textbf{64.30}&\cellcolor{gray!30} \textbf{79.64}&\cellcolor{gray!30} \textbf{76.16}&\cellcolor{gray!30} \textbf{50.87}&\cellcolor{gray!30} \textbf{54.14}&\cellcolor{gray!30} 65.28&\cellcolor{gray!30} \underline{83.70}&\cellcolor{gray!30} 48.86&\cellcolor{gray!30} 56.56&\cellcolor{gray!30} \textbf{83.26}&\cellcolor{gray!30} \textbf{75.23}&\cellcolor{gray!30} \underline{55.18}&\cellcolor{gray!30} \textbf{58.98}&\cellcolor{gray!30} \textbf{69.54}&\cellcolor{gray!30} 75.55&\cellcolor{gray!30} \textbf{44.33}&\cellcolor{gray!30} 49.43&\cellcolor{gray!30} 73.38\\ 
& Improv. & 9.28 & 9.39 & 11.74 & 10.87 & 5.30 & 4.67 & 8.14 & -0.76 & 1.29 & 2.24 & 1.48 & 2.05 & 8.08 & 3.09 & 8.80 & 11.08 & 2.40 & 2.17 & 5.74 & 1.57 \\

 \hline 
\multirow{5}{1cm}{\centering \texttt{GAT}~\cite{velivckovic2017graph}}  & \texttt{LLM-GNN} (FP) & 68.78 & 47.30 & 44.65 & 65.52&60.58&38.56&33.02&57.92&80.76&42.94&50.56&79.37&\underline{67.36}&48.87&47.07&\underline{62.73}&69.51&37.01&37.67&67.76 \\ 
& \texttt{LLM-GNN} (GP) &69.05&48.34&44.63&66.13&67.76&41.80&41.42&63.86&81.65&44.80&52.44&80.49&64.36&45.91&43.46&59.30&66.01&33.48&34.00&63.44\\ 
& \texttt{LLM-GNN} (RIM) &66.19&44.86&42.84&63.29&64.10&40.02&37.38&60.68&82.47&46.64&54.53&\underline{81.41}&66.85&47.51&46.12&61.97&68.76&38.08&36.80&67.46 \\  \cline{2-22}
& \algo &\cellcolor{gray!30} 76.06&\cellcolor{gray!30} 56.75&\cellcolor{gray!30} 55.78&\cellcolor{gray!30} 73.24&\cellcolor{gray!30} 73.97&\cellcolor{gray!30} 47.38&\cellcolor{gray!30} 50.27&\cellcolor{gray!30} \textbf{66.26}&\cellcolor{gray!30} 82.48&\cellcolor{gray!30} 46.08&\cellcolor{gray!30} 53.94&\cellcolor{gray!30} 81.94&\cellcolor{gray!30} 73.23&\cellcolor{gray!30} 54.22&\cellcolor{gray!30} 54.83&\cellcolor{gray!30} 67.80&\cellcolor{gray!30} \textbf{76.37}&\cellcolor{gray!30} 44.14&\cellcolor{gray!30} \textbf{50.75}&\cellcolor{gray!30} \textbf{74.12} \\ 
& Improv. & 7.01 & 8.41 & 11.13 & 7.11 & 6.21 & 5.58 & 8.85 & 2.40 & 0.01 & -0.56 & -0.59 & 0.53 & 5.87 & 5.35 & 7.76 & 5.07 & 6.86 & 6.06 & 13.08 & 6.36 \\  \hline
\multirow{5}{1cm}{\centering \texttt{GCNII}~\cite{chen2020simple}} & \texttt{LLM-GNN} (FP) & 69.37 & 49.12& 45.98 &65.73 &60.23&36.38&32.17&57.80&77.79&38.54&45.05&75.30&60.77&45.47&39.94&53.45&73.88&42.73&45.20&72.53 \\ 
& \texttt{LLM-GNN} (GP) &70.28&50.56&47.52&66.71&63.55&37.82&35.31&59.89&80.78&43.01&51.47&79.05&57.47&41.00&37.92&47.85&69.54&38.36&37.68&68.56 \\ 
& \texttt{LLM-GNN} (RIM) &69.52&49.01&47.75&65.85&65.04&40.26&38.48&61.63&79.81&42.47&50.60&77.14&64.90&46.22&43.44&59.08&66.72&39.13&33.92&66.41 \\  \cline{2-22}
& \algo &\cellcolor{gray!30} 78.71&\cellcolor{gray!30} 58.76&\cellcolor{gray!30} 58.45&\cellcolor{gray!30} 76.81&\cellcolor{gray!30} 73.14&\cellcolor{gray!30} 47.19&\cellcolor{gray!30} 49.28&\cellcolor{gray!30} 65.45&\cellcolor{gray!30} 82.08&\cellcolor{gray!30} 45.58&\cellcolor{gray!30} 53.44&\cellcolor{gray!30} 81.37&\cellcolor{gray!30} 68.70&\cellcolor{gray!30} 49.48&\cellcolor{gray!30} 50.53&\cellcolor{gray!30} 62.11&\cellcolor{gray!30} 75.23&\cellcolor{gray!30} 43.38&\cellcolor{gray!30} 48.60&\cellcolor{gray!30} 72.59\\ 
& Improv. & 8.43 & 8.20 & 10.70 & 10.10 & 8.10 & 6.93 & 10.80 & 3.82 & 1.30 & 2.57 & 1.97 & 2.32 & 3.80 & 3.26 & 7.09 & 3.03 & 1.35 & 0.65 & 3.40 & 0.06 \\ 
\hline

\end{tabular}
}
\end{small}
\label{tbl:quality}
\vspace{-2mm}
\end{table*}

\noindent
\textbf{Datasets and Metrics.} In this work, we use five benchmark TAG datasets for the node classification task, including \textit{Cora}, \textit{Citeseer}, \textit{Pubmed}, \textit{Wiki-CS}, and \textit{DBLP}. 
The statistics of these datasets are provided in Table~\ref{tab:datasets}. 
Following prior works~\cite{chen2023label}, we generate the attribute vector for each node using Sentence-BERT~\cite{reimers2019sentencebertsentenceembeddingsusing} as the text embedded to encode its associated text description.
Four widely-used metrics~\cite{liu2022deep}: {\em accuracy} (Acc), {\em normalized mutual information} (NMI), {\em adjusted Rand index} (ARI), and {\em F1-score} (F1), are adopted to assess the classification performance.

\noindent
\textbf{Baselines.} For a comprehensive comparison, we evaluate \algo against eight categories of baseline approaches using various model architectures or backbones.
The first methodology involves BERT-like architectures and we choose two {\em pretrained language models} (PLMs): BERT~\cite{devlin2019bertpretrainingdeepbidirectional}, and Sentence-BERT~\cite{reimers2019sentencebertsentenceembeddingsusing} with the {\em Sentence Embedding Similarity} metric. Additionally, we include BART-Large-MNLI~\cite{lewis2019bartdenoisingsequencetosequencepretraining}, a pretrained model fine-tuned on the MNLI dataset.
As for prompt engineering-based approaches, we go for two zero-shot prompt templates, i.e., zero-shot and zero-shot with Chain of Thought (COT), from \texttt{Graph-LLM}~\cite{chen2024exploring}.
As Label-free node classification is similar to Clustering, we also involve three clustering methods\footnote{More clustering baselines can be found in Appendix} as a comparison.
Similarly, we also include two {\em PLM+GNN} methods, OFA~\cite{liu2024alltraininggraphmodel} and ZeroG~\cite{li2024zeroginvestigatingcrossdatasetzeroshot}, that are also designed to work on zero-shot scenarios.
Furthermore, in the remaining three categories, we evaluate \algo with the state of the art, i.e., LLM-GNN~\cite{chen2023label}, with three popular GNN models (GCN~\cite{kipf2016semi}, GAT~\cite{velivckovic2017graph}, and GCNII~\cite{chen2020simple}) as the backbone, respectively.
Particularly, for each category, we consider three variants of LLM-GNN with FeatProp (FP)~\cite{wu2019active}, GraphPart (GP)~\cite{ma2022partition}, and RIM~\cite{zhang2021rim} as the active selection strategies.

\noindent
\textbf{Other Settings.} 
All experiments are conducted on a Linux machine powered by an Intel Xeon Platinum 8352Y CPU with 128 cores, 2TB of host memory, and four NVIDIA A800 GPUs, each with 80GB of device memory.
Unless otherwise specified, for the methods involving LLMs and GNNs, we employ \textit{GPT-3.5-turbo}~\cite{achiam2023gpt} as the LLM and adopt GCN~\cite{kipf2016semi} as the GNN backbone.
For each dataset, the numbers of annotations from LLMs are ensured to be the same for all evaluated approaches.
All reported results are averaged over three trials and each trial uses a different random seed for model training.
For the interest of space, more details regarding datasets, baselines, and hyperparameters are deferred to our supplementary material.

\subsection{RQ1. {Comparison with Zero-Shot Methods}}
Table~\ref{tbl:quality} presents the Acc, NMI, ARI, and F1 performance obtained by \algo and the aforementioned seven groups of baseline methods, i.e., 20 competitors, over the five datasets.
Note that the LLM outputs by \texttt{Graph-LLM}~\cite{chen2024exploring} are invalid for F1-score calculation, and thus, are omitted.
From Table~\ref{tbl:quality}, we can make the following observations.
\begin{enumerate}[leftmargin=*,label=(\arabic*)] 
\item \algo achieves state-of-the-art results compared to the runner-up method, LLM-GNN. For instance, on \textit{Cora}, \textit{CitSeer}, and \textit{WikiCS}, using GCN as the backbone, \algo demonstrates significant improvements of 9.28\%, 5.30\%, and 8.08\%, respectively, in terms of classification accuracy. 
On \textit{Pubmed} and \textit{DBLP}, \algo attains smaller yet still a notable margin of 1.29\% and 2.40\% in Acc, respectively. 
When using GAT as the backbone, \algo still outperforms the best LLM-GNN counterpart with a considerable accuracy gain of 7.01\%, 6.21\%, 5.81\%, and 6.86\% on \textit{Cora}, \textit{Citeseer}, \textit{WikiCS}, and \textit{DBLP}, respectively. 
With GCNII, the improvements on \textit{Cora} and \textit{Citeseer} achieved by \algo are up to 8.43\% and 8.10\%, respectively. Similar observations can be made for other metrics.
These results underscore the superior performance of \algo and its versatile adaptability across various GNN models, which further validate the cost-effectiveness of our proposed techniques in combining GNNs and LLMs.
\item Moreover, among BERT-based PLMs, Sentence-BERT consistently achieves the best performance, with substantial accuracy improvements of 26.84\%, 20.18\%, 27.85\%, and 31.66\% on \textit{Cora}, \textit{Citeseer}, \textit{Pubmed}, and \textit{DBLP}, respectively.
This implies its effectiveness in encoding textual attributes in TAGs. However, without downstream fine-tuning, it remains significantly inferior to GNN-based methods.
\item \texttt{PLM+GNN} methods, including OFA~\cite{liu2024alltraininggraphmodel} and ZeroG~\cite{li2024zeroginvestigatingcrossdatasetzeroshot}, are designed to achieve cross-dataset zero-shot transferability by fine-tuning PLMs on similar semantic space. This approach represents a fundamentally different paradigm compared to our methodology. Consequently, directly adapting these methods to our label-free setting poses considerable challenges. In contrast, \algo demonstrates superior performance under this setting, highlighting its robustness and effectiveness.
\item Compared to \texttt{LLM as Predictor} method, which directly employs LLMs for annotation and label inference, \algo dominates its two versions with approximately 10\% gains in Acc, on almost all datasets.
This finding manifests the efficacy of our proposed graph-based techniques in exploiting the structural semantics underlying TAGs for node classification.
Note that the only exception is {\em PubMed}, where \texttt{Graph-LLM} yields superior performance over \algo due to its high-quality textual contents.
\end{enumerate}

\begin{table}[!t]
\centering
\caption{Ablation study of \algo.}
\renewcommand{\arraystretch}{1.0}
\label{tab:ablation}
\vspace{-2ex}
\resizebox{\columnwidth}{!}{%
\begin{tabular}{c|c|c|c|c|c}
\hline
 &  \textbf{\em Cora} & \textbf{\em Citeseer} & \textbf{\em PubMed} & \textbf{\em WikiCS} & \textbf{\em DBLP} \\
\hline
\algo & \textbf{81.80} & \textbf{76.16} & \textbf{83.70} & \textbf{75.23} & \textbf{75.55} \\ \hline
{w/o Initial Active Node Selection} & $77.67$ & $ 73.55$ & $81.70$ &  $65.49$ & $74.76$ \\
{w/o Informative Sample Selection} & {78.72} & {74.26} & {81.74} & {74.41} & {75.03} \\
{w/o Hybrid Label Refinement} & $78.65$ & $73.82$ & $ 82.55$ & $74.15$ & $68.95$ \\
\hline
\end{tabular}
}
\vspace{0ex}
\end{table}

\begin{table}[!t]
\centering
\caption{Ablation Study on Active Node Selection in \algo }
\label{tab:kmeans}
\vspace{-2ex}
\begin{tabular}{c|c|c|c}
\hline
 & {\algo ($K$-Means)} & \algo & Improv. \\
\hline
\bf{ {\em Cora}}& 73.83 & 81.80 & 7.97 \\ 
\bf{ {\em Citeseer}}& 67.05 & 76.16& 9.11\\ 
\bf{ {\em Pubmed}}& 71.58 & 83.70 &12.12 \\ 
\bf{ {\em WikiCS}}& 65.26 & 75.23&9.97\\
\bf{ {\em DBLP}}& 73.21 & 75.55 &2.34\\
\hline
\end{tabular}
\vspace{-2ex}
\end{table}

\subsection{RQ2. {Ablation Studies of \algo}}
\label{sec:ablation}
\stitle{Analysis of each module in \algo}
This section studies the effectiveness of each of the three major modular ingredients in \algo, i.e., {\em initial active node selection} in Section~\ref{sec:initial}, {\em informative sample selection} in Section~\ref{sec:sample-selection}, and the {\em hybrid label refinement} in Section~\ref{sec:hybrid}.
To be precise, we create three variants of \algo by replacing the active node selection strategy with the random selection, simply using the predictive probabilities in Eq.~\eqref{eq:Yr} to construct $\VCS$ and $\VUS$, and disabling the label refinement component, respectively.

As reported in Table~\ref{tab:ablation}, \algo consistently outperforms all these variants on all datasets in classification accuracy. On the tested datasets, there is a marked performance gap between each of the three variants and \algo, which exhibit the efficacy of our proposed techniques in \algo.
In particular, the variant, \algo{} {w/o \em initial active node selection}, produces the lowest accuracy results, indicating the importance of selecting representative nodes as the initial training samples.

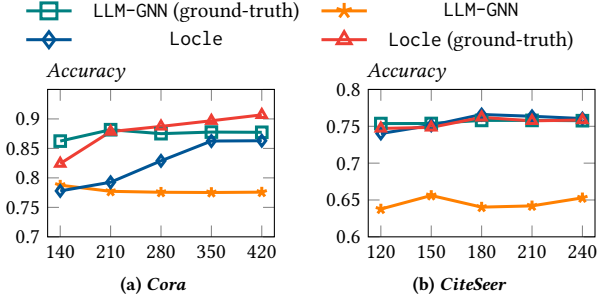
\begin{figure}[!t]
\centering
\begin{small}
\begin{tikzpicture}
    \begin{customlegend}
    [legend columns=2,
        legend entries={\texttt{LLM-GNN} (ground-truth), \texttt{LLM-GNN}, \algo, \algo (ground-truth)},
        legend style={at={(0.45,1.35)},anchor=north,draw=none,font=\small,column sep=0.2cm}]
    \addlegendimage{line width=0.4mm,mark size=3pt,mark=square,color=teal}
    \addlegendimage{line width=0.4mm,mark size=3pt,mark=star,color=orange}
    \addlegendimage{line width=0.4mm,mark size=3pt,mark=diamond,color=ACMDarkBlue}
        \addlegendimage{line width=0.4mm,mark size=3pt,mark=triangle,color=ACMRed}
    \end{customlegend}
\end{tikzpicture}
\\[-\lineskip]
\vspace{-4mm}
\subfloat[\em Cora]{
\begin{tikzpicture}[scale=1,every mark/.append style={mark size=2pt}]
    \begin{axis}[
        height=\columnwidth/2.4,
        width=\columnwidth/1.85,  
        ylabel={\it Accuracy},
        xmin=0.5, xmax=9.5,
        ymin=0.65, ymax=0.9,
        xtick={1,3,5,7,9},
        ytick={0.6,0.65,0.7,0.75,0.8,0.85,0.9},
        xticklabel style = {font=\small},
        yticklabel style = {font=\small},
        xticklabels={140,210,280,350,420},
        yticklabels={0.65,0.7,0.75,0.8,0.85,0.9},
        every axis y label/.style={font=\small,at={(current axis.north west)},right=5mm,above=0mm},
        legend style={fill=none,font=\small,at={(0.02,0.99)},anchor=north west,draw=none},
    ]
    \addplot[line width=0.4mm, mark=square,color=teal]  
        plot coordinates {
(1,	0.8124	)
(3,	0.8317	)
(5,	0.8250	)
(7,	0.8277	)
(9, 0.8273)
};

    \addplot[line width=0.4mm, mark=star,color=orange]  
        plot coordinates {
(1,	0.7376	)
(3,	0.7272	)
(5,	0.7256	)
(7, 0.7252)
(9, 0.7258)
    };
    \addplot[line width=0.4mm, mark=diamond,color=ACMDarkBlue]  
        plot coordinates {
(1,	0.7279	)
(3,	0.7426	)
(5,	0.7789	)
(7, 0.8125)
(9, 0.8130)
    };
        \addplot[line width=0.4mm, mark=triangle,color=ACMRed]  
        plot coordinates {
(1,	0.7742	)
(3,	0.8285	)
(5,	0.8373	)
(7, 0.8469)
(9, 0.8569)
    };
    \end{axis}
\end{tikzpicture}\hspace{4mm}\label{fig:B-cora}%
}
\subfloat[\em CiteSeer]{
\begin{tikzpicture}[scale=1,every mark/.append style={mark size=2pt}]
    \begin{axis}[
        height=\columnwidth/2.4,
        width=\columnwidth/1.85,  
        ylabel={\it Accuracy},
        xmin=0.5, xmax=9.5,
        ymin=0.6, ymax=0.8,
        xtick={1,3,5,7,9},
        ytick={0.6,0.65,0.7,0.75,0.8},
        xticklabel style = {font=\small},
        yticklabel style = {font=\small},
        xticklabels={120,150 ,180, 210, 240, },
        yticklabels={0.6,0.65,0.7,0.75,0.8},
        every axis y label/.style={font=\small,at={(current axis.north west)},right=5mm,above=0mm},
        legend style={fill=none,font=\small,at={(0.02,0.99)},anchor=north west,draw=none},
    ]
    \addplot[line width=0.4mm, mark=square,color=teal]  
        plot coordinates {
(1,	0.7537	)
(3,	0.7538	)
(5,	0.7580	)
(7,	0.7580	)
(9,	0.7576	)
    };

   \addplot[line width=0.4mm, mark=star,color=orange]  
        plot coordinates {
(1,	0.6377 )
(3,	0.6561	)
(5,	0.6403	)
(7,	0.6420	)
(9,	0.6529	)
};
   \addplot[line width=0.4mm, mark=diamond,color=ACMDarkBlue]  
        plot coordinates {
(1,	0.7402 )
(3,	0.7513	)
(5,	0.7660	)
(7,	0.7636	)
(9,	0.7605	)
};
  \addplot[line width=0.4mm, mark=triangle,color=ACMRed]  
        plot coordinates {
(1,	0.7471	)
(3,	0.7485	)
(5,	0.7621	)
(7,	0.7574	)
(9, 0.7585)
    };
    \end{axis}
\end{tikzpicture}\hspace{0mm}\label{fig:B-citeseer}%
}
\end{small}
 \vspace{-3mm}
\caption{Varying $B$ in \algo.} \label{fig:B_analysis}
\vspace{-3ex}
\end{figure}

\stitle{Analysis of different Active Node Selection methods}
To verify the efficacy of our subspace clustering technique for active node selection in Section~\ref{sec:selection}, we substitute $K$-Means for it in \algo. As reported in Table~\ref{tab:kmeans}, the variant of \algo with $K$-Mean consistently exhibits remarkable performance degradation compared to the version using the subspace clustering. For example, on the {\em Pubmed} dataset, the subspace clustering approach attains a large margin of $12.12\%$ in terms of classification accuracy compared to $K$-Means. The reason is that $K$-Means tends to group the majority of nodes into only a few clusters, making it hard to select sufficient representative nodes for annotation. In contrast, as elaborated in Section~\ref{sec:selection}, our subspace clustering approach can filter out the noise and identify the inherent structure underlying the node features for a more balanced partition.

\subsection{RQ3. {Hyperparameter Analysis of \algo}}

\stitle{Analysis of the Budget Size $B$}
Recall that in Section~\ref{sec:intro} (see Figure~\ref{fig:example}), a severe limitation of \texttt{LLM-GNN}~\cite{chen2023label} is that increasing the query budget $B$ does not always lead to performance improvements. Additionally, there is a large gap between the performance achieved by LLM-annotated labels and that obtained with ground-truth labels. 
As shown in Figures~\ref{fig:B-cora} and~\ref{fig:B-citeseer}, we report the accuracy performance of \algo, \texttt{LLM-GNN} and their variants using ground-truth labels (dubbed as \algo (ground-truth) and \texttt{LLM-GNN} (ground-truth), respectively) when increasing the query budget $B$ from $140$ to $420$, and from $120$ to $240$, on {\em Cora} and {\em CiteSeer}, respectively. 
It can be observed that the performance of \algo steadily improves as $B$ increases. Notably, when $B\ge 350$, \algo achieves comparable performance to methods using ground-truth labels on the {\em Cora} dataset. For the {\em CiteSeer} dataset, \algo remains highly competitive across all budget settings. These findings demonstrate the effectiveness of our proposed techniques in utilizing GNNs for active node selection, node annotation, and label correction.


\begin{figure}[!t]
\centering

\begin{tikzpicture}
    \begin{customlegend}
    [legend columns=3,
        legend entries={{\em Cora}, {\em Citeseer}, {\em Pubmed}, {\em WikiCS}},
        legend style={at={(0.45,1.35)},anchor=north,draw=none,font=\small,column sep=0.2cm}]
    \addlegendimage{line width=0.4mm,mark size=3pt,mark=triangle,color=teal}
    \addlegendimage{line width=0.4mm,mark size=3pt,mark=diamond,color=orange}
    \addlegendimage{line width=0.4mm,mark size=3pt,mark=square,color=ACMDarkBlue}
    \end{customlegend}
    \begin{customlegend}
    [legend columns=2,
        legend entries={{\em WikiCS}, {\em DBLP}},
        legend style={at={(0.45,0.95)},anchor=north,draw=none,font=\small,column sep=0.2cm}]
    \addlegendimage{line width=0.4mm,mark size=3pt,mark=star,color=ACMRed}
    \addlegendimage{line width=0.4mm,mark size=3pt,mark=pentagon,color=ACMPurple}
    \end{customlegend}
\end{tikzpicture}
\vspace{-4mm}

\subfloat[Varying $\varepsilon$ in \algo]{
\begin{tikzpicture}[scale=1,every mark/.append style={mark size=2pt}]
    \begin{axis}[
        height=\columnwidth/2.3,
        width=\columnwidth/1.85,  
        ylabel={\it Accuracy},
        xmin=0.5, xmax=9.5,
        ymin=0.65, ymax=0.85,
        xtick={1,3,5,7,9},
        ytick={0.65,0.7,0.75,0.8,0.85},
        xticklabel style = {font=\small},
        yticklabel style = {font=\small},
        xticklabels={0.1,0.3,0.5,0.7,0.9,},
        yticklabels={0.65,0.7,0.75,0.8,0.85},
        every axis y label/.style={font=\small,at={(current axis.north west)},right=5mm,above=0mm},
        legend style={fill=none,font=\small,at={(0.02,0.99)},anchor=north west,draw=none},
    ]
    \addplot[line width=0.4mm, mark=triangle,color=teal]  
        plot coordinates {
(1,	0.7698	)
(3,	0.8037	)
(5,	0.8180	)
(7,	0.7927	)
(9,	0.7500	)
    };
    \addplot[line width=0.4mm, mark=diamond,color=orange]  
        plot coordinates {
(1,	0.7378	)
(3,	0.7392	)
(5,	0.7605	)
(7,	0.7525	)
(9,	0.7282	)
    };
    \addplot[line width=0.4mm, mark=square,color=ACMDarkBlue]  
        plot coordinates {
(1,	0.6821	)
(3,	0.7986	)
(5,	0.8370	)
(7,	0.8090	)
(9,	0.8139	)
    };
    \addplot[line width=0.4mm, mark=star,color=ACMRed]  
        plot coordinates {
(1,	0.7419	)
(3,	0.7144	)
(5,	0.7503	)
(7,	0.7468	)
(9,	0.7320	)
    };
    \addplot[line width=0.4mm, mark=pentagon,color=ACMPurple]  
        plot coordinates {
(1,	0.6634	)
(3,	0.7064	)
(5,	0.7129	)
(7,	0.7318	)
(9,	0.7226	)
    };
    \end{axis}
\end{tikzpicture}\hspace{4mm}\label{fig:tau_analysis}}%
\subfloat[Varying $R$ in \algo]{
\begin{tikzpicture}[scale=1,every mark/.append style={mark size=2pt}]
    \begin{axis}[
        height=\columnwidth/2.3,
        width=\columnwidth/1.85,  
        ylabel={\it Accuracy},
        xmin=0.5, xmax=9.5,
        ymin=0.7, ymax=0.85,
        xtick={1,3,5,7,9},
        ytick={0.7,0.75,0.8,0.85},
        xticklabel style = {font=\small},
        yticklabel style = {font=\small},
        xticklabels={3,4,5,6,7},
        yticklabels={0.7,0.75,0.8,0.85},
        every axis y label/.style={font=\small,at={(current axis.north west)},right=5mm,above=0mm},
        legend style={fill=none,font=\small,at={(0.02,0.99)},anchor=north west,draw=none},
    ]
    \addplot[line width=0.4mm, mark=triangle,color=teal]  
        plot coordinates {
(1,	0.8001	)
(3,	0.8086	)
(5,	0.8181	)
(7, 0.7919)
(9, 0.7954)
    };
    \addplot[line width=0.4mm, mark=diamond,color=orange]  
        plot coordinates {
(1,	0.7469	)
(3,	0.7586	)
(5,	0.7621	)
(7, 0.7527)
(9, 0.7482)
    };
    \addplot[line width=0.4mm, mark=square,color=ACMDarkBlue]  
        plot coordinates {
(1,	0.8210	)
(3,	0.8229	)
(5,	0.8370	)
(7, 0.8227)
(9, 0.8227)
    };
    \addplot[line width=0.4mm, mark=star,color=ACMRed]  
        plot coordinates {
(1,	0.7327	)
(3,	0.7425	)
(5,	0.7320	)
(7, 0.7473)
(9,0.7408)
    };
    \addplot[line width=0.4mm, mark=pentagon,color=ACMPurple]  
        plot coordinates {
(1,	0.7384	)
(3,	0.7411	)
(5,	0.7495	)
(7, 0.7666)
(9,0.7521)
    };
    \end{axis}
\end{tikzpicture}\hspace{6mm}\label{fig:R_analysis}}%
\vspace{-3mm}
\caption{Varying $R$ and $\varepsilon$ in \algo.} \label{fig:allocation}
\vspace{-3ex}
\end{figure}
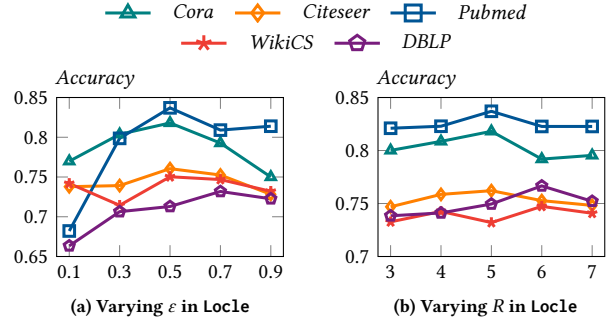

\stitle{Analysis of Budge Allocation Ratio $\varepsilon$}
We further investigate how the budget allocation ratio $\varepsilon$ affects the performance of \algo, which determines the number of annotations used in the initial stage and subsequent self-training stages.
Figure~\ref{fig:tau_analysis} shows Acc results across different $\varepsilon$ values (0.1-0.9) on five datasets. Results indicate $\varepsilon$ significantly impacts model performance, with $\varepsilon=0.5$ typically yielding best results. When $\varepsilon$ is too small, some categories lack sufficient labeled samples for effective prediction. Conversely, large $\varepsilon$ values reduce annotations for self-training, limiting the model's ability to leverage graph structures for refinement and making it overly dependent on LLM-generated labels.


\begin{table}[!t]
\centering
\caption{Runtime Cost (sec) Analysis of \algo }
\label{tab:time}
\vspace{-2ex}
\begin{tabular}{c|c|c|c|c}
\hline
 & {\texttt{LLM-GNN}} & {\texttt{LLM-GNN}} & {\texttt{LLM-GNN}} & \multirow{2}{*}{\centering {\algo}} \\
 & (\texttt{FeatProp}) & (\texttt{Graphpart}) & (\texttt{RIM}) &  \\
\hline
\bf{ {\em Cora}}& 26.8 & 40.4 & 141.9 &50.9 \\ 
\hline
\bf{ {\em Pubmed}}& 39.4 & 53.2 &776.8&217.6\\ 
\hline
\bf{ {\em WikiCS}}& 52.2 &60.1 &934.7&217.1\\
\hline
\bf{ {\em DBLP}}& 72.5 &46.5&500.2&208.7\\
\hline
\end{tabular}
\vspace{-3ex}
\end{table}

\begin{table}[!t]
\centering
\caption{Monetary Cost (US\$) Analysis of \algo and entire dataset}
\label{tab:cost}
\vspace{-2ex}
\begin{tabular}{c|c|c|c|c|c}
\hline
 & \textbf{\em Cora} & \textbf{ {\em Citeseer}} &  \textbf{ {\em Pubmed}} & \textbf{ {\em WikiCS}} & \textbf{ {\em DBLP}} \\
\hline
\algo{} & 0.10 & 0.07 & 0.09 & 0.35 & 0.01 \\
{Entire} & 1.33 & 1.81 & 21.83 & 18.65 & 0.50 \\

\hline
\end{tabular}
\vspace{-4ex}
\end{table}

\stitle{Analysis of the Training Round $R$}
We empirically study how the number of self-training rounds $R$ affects \algo's performance.
Figure~\ref{fig:R_analysis} shows classification accuracy as $R$ increases from 3 to 7.
Across all datasets, performance first rises then slightly declines with increasing $R$. Smaller TAGs ({\em Cora}, {\em CiteSeer}, {\em PubMed}) perform best at $R=5$, while larger datasets ({\em WikiCS}, {\em DBLP}) achieve optimal results at $R=6$.
With fixed budget $B$, query budget per round equals $\frac{B-B_{ini}}{R-1}$, meaning higher $R$ values result in fewer newly annotated samples per round, potentially introducing label noise that limits performance gains.

\subsection{RQ4. {Cost Analysis of \algo}}
In this section, we first analyze the training time of various LLM-GNN~\cite{chen2023label} variants and \algo, as presented in Table~\ref{tab:time}. As shown, \algo is at least $2\times$ faster than the best-performing variant of the state-of-the-art LLM-GNN, namely LLM-GNN (RIM). While LLM-GNN (Featprop) and LLM-GNN (Graphpart) adopt faster active node selection strategies, they sacrifice accuracy for speed. In contrast, although \algo introduces additional training steps, its active node selection approach, as described in Sec.~\ref{sec:selection}, achieves high effectiveness with relatively low computational cost. In comparison, the label selection process in LLM-GNN (RIM) relies on an iterative batch setting, which is more computationally expensive.

Moreover, table~\ref{tab:cost} compares the financial costs of querying LLMs in \algo versus annotating entire datasets, based on GPT-3.5-turbo's tokenizer and OpenAI's pricing.

\section{Conclusion}
In this work, we present \algo, a cost-effective solution that integrates LLMs into GNNs for label-free node classification. 
\algo achieves high result utility through three major contributions:
(i) an effective active node selection strategy for initial annotation via LLMs, (ii) a sample selection scheme that accurately identifies informative nodes based on our proposed label disharmonicity and entropy, and (iii) a label refinement module that combines the strengths of LLMs and GNNs with a rewired graph topology. 
Our extensive experiments over 5 real-world TAG datasets demonstrate the superiority of \algo over the state-of-the-art methods.
In the future, we intend to scale \algo to large TAGs and extend it to other graph-related tasks or information retrieval applications, such as link prediction, graph classification, document categorization, and item tagging.


\begin{acks}
This work was supported by National Key Research and Development Program (Grant No. 2022YFB4501400), the Hong Kong RGC ECS grant (No. 22202623), the National Natural Science Foundation of China (Grant No. 62302414, 62202451), CAS Project for Young Scientists in Basic Research (Grant No. YSBR-029), CAS Project for Youth Innovation Promotion Association, and the Huawei Gift Fund.

\end{acks}

\balance

\bibliographystyle{ACM-Reference-Format}
\bibliography{sample-base}

\clearpage
\newpage
\appendix
{
\section{Theoretical Proofs}\label{sec:proof}

\begin{proof}[\bf Proof of Lemma~\ref{lem:sc-kmeans}]
Let $\{\mathcal{C}_1,\ldots,\mathcal{C}_K\}$ be the $K$ clusters and $\CM\in \mathbb{R}^{|\V|\times K}$ be a node-cluster indicator wherein $\CM_{i,k}=\frac{1}{\sqrt{|\mathcal{C}_k|}}$ if $v_i\in \mathcal{C}_k$, and $0$ otherwise. Recall that the spectral clustering of affinity graph $\frac{\SM+\SM^{\top}}{2}=\SM$ is to find $\CM$ such that the following objective is optimized:
\begin{equation}\label{eq:sc-obj}
\max_{\CM}\Tr(\CM^{\top}\SM\CM).
\end{equation}

Next, recall that $K$-Means seeks to minimize the Euclidean distance between data points and their respective cluster centroids, which leads to
\begin{small}
\begin{align*}
& \min_{\CM} \sum_k^K \sum_{v_i \in \mathcal{C}_k} \left\Vert \UM_i - \frac{1}{\vert \mathcal{C}_k\vert}\sum_{v_j \in \mathcal{C}_k}{\UM_j} \right\Vert_2^2 \\
= & \sum_k^K \sum_{v_i \in \mathcal{C}_k} \Vert \UM_i \Vert_2^2 - \sum_k^K \sum_{v_j, v_\ell \in \mathcal{C}_k}{\frac{\UM_j\cdot \UM_\ell^{\top}}{|\mathcal{C}_k|}}\\
= & \sum_k^K \sum_{v_i \in \mathcal{C}_k} \Vert \UM_i \Vert_2^2 - \Tr(\CM^{\top}\UM\UM^{\top}\CM).
\end{align*}
\end{small}
Since the first term $\sum_k^K \sum_{v_i \in \mathcal{C}_k} \Vert \UM_i \Vert_2^2$ is fixed, the above optimization objective is equivalent to maximizing $\Tr(\CM^{\top}\UM\UM^{\top}\CM)$, which finishes the proof as $\SM=\UM\UM^{\top}$.
\end{proof}

\begin{proof}[\bf Proof of Lemma~\ref{lem:laplacian}]
Recall that $\tilde{\LM}=\IM-\NAM$ in Section~\ref{sec:problem}. Accordingly, the objective of $\min_{\tilde{\LM}}{\Tr(\HM^{\top}\tilde{\LM}\HM)}$ is equivalent to optimizing $\max_{\NAM}{\Tr(\HM^{\top}\NAM\HM)}$.

Next, we rewrite $\Tr(\HM^{\top}\NAM\HM)$ as $\Tr(\NAM\HM\HM^{\top})$ using the cyclic property of the trace. Based on the definition of matrix trace and Cauchy-Schwarz inequality,
\begin{align*}
\Tr(\NAM\HM\HM^{\top}) & = \sum_{i,j=1}^n{\NAM_{i,j} (\HM\HM^{\top})_{j,i}}  \le \sqrt{\sum_{i,j=1}^n{\NAM_{i,j}^2}} \sqrt{\sum_{i,j=1}^n{(\HM\HM^{\top})_{j,i}^2}} \\
& = \|\NAM\|_F \cdot \sqrt{\sum_{i,j=1}^n{(\HM\HM^{\top})_{j,i}^2}} = {\beta} \cdot \|\HM\HM^\top\|_F.
\end{align*}
The maximum can be achieved when $\NAM_{j,i}=\frac{\beta\cdot(\HM\HM^{\top})_{j,i}}{\|\HM\HM^\top\|_F}$, which completes the proof.
\end{proof}

\section{Experimental Details}\label{sec:exp-detail}

\subsection{Details of the Prompt}\label{sec:prompt}
In this section, we present the prompts designed for annotation with two examples. For \textit{Cora}, \textit{Citeseer}, \textit{Pubmed}, and \textit{WikiCS}, we incorporate label reasoning. Given that the reasoning text adds only a minimal cost compared to the longer query text in these datasets, its inclusion is acceptable and can slightly improve performance. However, for \textit{DBLP}, where the query text is short, we did not include additional reasoning text. Tables ~\ref{prompt:0_shot_pubmed} and ~\ref{prompt:0_shot_dblp} show the complete structure of our prompts. Specially, for \textit{WikiCS}, we truncate the text of the input article and make sure the total prompt is less than 4096 tokens.
\begin{table}[H]
\caption{Full prompt example for zero-shot annotation with a confidence score for \textit{Pubmed}}
\label{prompt:0_shot_pubmed}
\centering
\begin{tabularx}{\linewidth}{X}  
\toprule
\textbf{Input:} \\
You are a model that is especially good at classifying a paper's category. Now I will first give you all the possible categories and their explanations. Please answer the following question: What is the category of the target paper? \\
\textbf{All possible categories:} [Diabetes Mellitus Experimental, Diabetes Mellitus Type 1, Diabetes Mellitus Type 2] \\
\textbf{Category explanation:} 
Diabetes Mellitus Experimental: General experimental studies on diabetes, including comparisons across different animal models. 

Diabetes Mellitus Type 1: Studies specific to Type 1 diabetes, involving metabolic changes and skeletal muscle metabolism. 

Diabetes Mellitus Type 2: Studies specific to Type 2 diabetes, often involving complications, metabolic studies, and preventive measures.

\textbf{Target paper}:

\textbf{Title}: Isolated hyperglycemia at 1 hour on oral glucose tolerance test in pregnancy resembles gestational diabetes mellitus in predicting postpartum metabolic dysfunction.

\textbf{Abstract}: OBJECTIVE: Gestational impaired glucose tolerance (GIGT), defined by a single abnormal value on antepartum 3-h oral glucose tolerance test (OGTT) ... CONCLUSIONS: Like GDM, 1-h GIGT is associated with postpartum glycemia, insulin resistance, and beta-cell dysfunction. \\
Output your answer together with a confidence score ranging from 0 to 100, in the form of a list of Python dicts like \texttt{[{"answer": <answer\_here>, "confidence": <confidence\_here>}]}.
You only need to output the one answer you think is the most likely. \\
\textbf{Output:} \\
\bottomrule
\end{tabularx}
\end{table}

\begin{table}[H]
\caption{Full prompt example for zero-shot annotation with a confidence score for \textit{DBLP}}
\label{prompt:0_shot_dblp}
\centering
\begin{tabularx}{\linewidth}{X}  
\toprule
\textbf{Input:} \\
You are a model that is especially good at classifying a paper's category. Now I will first give you all the possible categories and their explanations. Please answer the following question: What is the category of the target paper? \\
\textbf{All possible categories:} [Database, Data Mining, AI, Information Retrieval] \\
\textbf{Target Paper:} 

\textbf{Title:} Panel: User Modeling and User Interfaces. \\
Output your answer together with a confidence score ranging from 0 to 100, in the form of a list of Python dicts like \texttt{[{"answer": <answer\_here>, "confidence": <confidence\_here>}]}.
You only need to output the one answer you think is the most likely. \\
\textbf{Output:} \\
\bottomrule
\end{tabularx}
\end{table}


\begin{table*}[!t]
\centering
\caption{Dataset descriptions.}
\vspace{-2ex}
\label{tab:stats_full}
\begin{small}
\begin{tabular}{c|c|c|m{4.5cm}|m{6cm} }
\hline
{\bf Dataset} & {\bf \#Nodes} & {\bf \#Edges} & \multicolumn{1}{c|}{\bf Task Description} & \multicolumn{1}{c}{\bf Classes } \\ \hline
{\em Cora} & $2,708$ & $5,429$ & Given the title and abstract,  predict the category of this paper & Rule Learning, Neural Networks, Case Based, 
Genetic Algorithms, Theory, Reinforcement Learning, Probabilistic Methods \\ \hline
{\em Citeseer} & $3,186$ & $4,277$ & Given the title and abstract, predict the category of this paper & Agents, Machine Learning, Information Retrieval, Database, Human Computer Interaction, Artificial Intelligence \\ \hline
{\em Pubmed} & $19,717$ & $44,335$ & Given the title and abstract,  predict the category of this paper & Diabetes Mellitus Experimental, Diabetes Mellitus Type 1, Diabetes Mellitus Type 2 \\ \hline
{\em WikiCS} & $11,701$ & $215,863$ & Given the contents of  the Wikipedia article,  predict the category of this  article & Computational linguistics, Databases, Operating systems, Computer architecture, Computer security, Internet protocols, Computer file systems, Distributed computing architecture,  Web technology, Programming language topics \\ \hline
{\em DBLP} & $14,376$ & $431,326$ & Given the title of the paper in dblp database, predict the category of the paper & Database, Data Mining,  AI, Information Retrieval \\ \hline
\end{tabular}
\end{small}
\end{table*}

\begin{table*}[ht]
\centering
\caption{Hyperparameters for \algo with GCN, GAT, and GCNII backbones.}
\renewcommand{\arraystretch}{0.8}
\vspace{-2ex}
\resizebox{\textwidth}{!}{%
\begin{tabular}{p{1.5cm} p{0.6cm} p{0.7cm} p{0.7cm} p{0.8cm} p{0.7cm} | p{0.6cm} p{0.7cm} p{0.7cm} p{0.8cm} p{0.7cm} | p{0.6cm} p{0.7cm} p{0.7cm} p{0.8cm} p{0.7cm}}
\toprule
 & \multicolumn{5}{c}{\textbf{GCN}} & \multicolumn{5}{c}{\textbf{GAT}} & \multicolumn{5}{c}{\textbf{GCNII}} \\
\cmidrule(lr){2-6} \cmidrule(lr){7-11} \cmidrule(lr){12-16}
 & {\em Cora} & {\em Citeseer} & {\em Pubmed} & {\em WikiCS} & {\em DBLP} 
 & {\em Cora} & {\em Citeseer} & {\em Pubmed} & {\em WikiCS} & {\em DBLP}
 & {\em Cora} & {\em Citeseer} & {\em Pubmed} & {\em WikiCS} & {\em DBLP}  \\
\midrule
lr & 0.01 & 0.01 & 0.01 & 0.01 & 0.1 & 0.01 & 0.01 & 0.01 & 0.01 & 0.1 & 0.01 & 0.01 & 0.01 & 0.01 & 0.05 \\
dropout & 0.5 & 0.5 & 0.5 & 0.5 & 0.5 & 0.5 & 0.5 & 0.2 & 0.5 & 0.5 & 0.5 & 0.5 & 0.5 & 0.5 & 0.5 \\
\#layers & 2 & 2 & 2 & 2 & 2 & 2 & 2 & 2 & 2 & 2 & 64 & 32 & 16 & 8 & 2 \\
hidden dim. & 64 & 64 & 64 & 64 & 64 & 64 & 64 & 16 & 64 & 64 & 64 & 64 & 64 & 64 & 64 \\
$T$ & 2 & 2 & 2 & 2 & 2 & 2 & 2 & 2 & 2 & 2 & 2 & 2 & 2 & 2 & 2 \\
$\alpha$ & 1 & 1 & 1.2 & 1 & 1 & 1 & 1 & 1.2 & 1 & 1 & 1 & 1 & 1.2 & 1 & 1 \\
$\epsilon$ & 0.5 & 0.4 & 0.5 & 0.4 & 0.25 & 0.4 & 0.4 & 0.5 & 0.3 & 0.4 & 0.4 & 0.33 & 0.5 & 0.4 & 0.3 \\
$\tau$ & 128 & 128 & 256 & 256 & 64 & 128 & 128 & 4 & 128 & 32 & 128 & 256 & 4 & 128 & 128 \\
$\lambda$ & 5e-5 & 1e-4 & 5e-3 & 5e-3 & 5e-3 & 5e-3 & 1e-4 & 5e-3 & 5e-3 & 5e-3 & 5e-3 & 1e-4 & 5e-3 & 5e-3 & 5e-3\\
$B$ & 350 & 200 & 150 & 400 & 320 & 350 & 200 & 150 & 400 & 240 & 350 &200 & 150& 400&240 \\
$\delta^{(-)}$ & 0.1 & 0 & 0 & 0.01 & 0 & 0.1 &0 &0 &0 &0.1 & 0&0 &0.05 &0.1 &0.01 \\
$\delta^{(+)}$ & 0.05 & 0.02 & 0 & 0.3 & 0.05 & 0.05& 0.01&0.3 &0.01 &0 &0.01 &0.1 &0.1 &0.01 &0 \\
$\overline{\phi}$ & 3 & 3 & 5 & 3 & 3 & 3 &3 &5 &3 & 3&3 & 3& 3& 3& 3\\
\#epochs & 20 & 15 & 30 & 30 & 20 & 20&15 &30 &30 &30 & 100&100 &100 &100 &20 \\
\bottomrule
\end{tabular}
}
\label{tbl:combined_param_selection}
\end{table*}

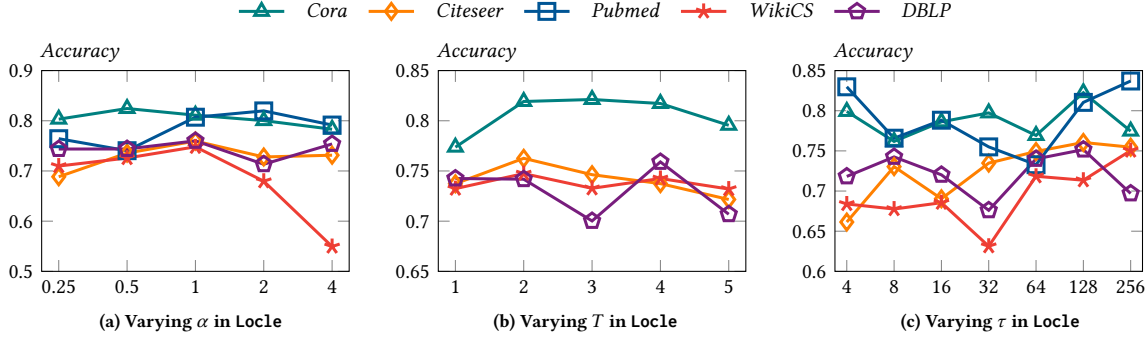
\begin{figure*}[!t]
\centering
\begin{tikzpicture}
    \begin{customlegend}
    [legend columns=5,
        legend entries={{\em Cora}, {\em Citeseer}, {\em Pubmed}, {\em WikiCS}, {\em DBLP}},
        legend style={at={(0.45,1.35)},anchor=north,draw=none,font=\small,column sep=0.2cm}]
    \addlegendimage{line width=0.4mm,mark size=3pt,mark=triangle,color=teal}
    \addlegendimage{line width=0.4mm,mark size=3pt,mark=diamond,color=orange}
    \addlegendimage{line width=0.4mm,mark size=3pt,mark=square,color=ACMDarkBlue}
    \addlegendimage{line width=0.4mm,mark size=3pt,mark=star,color=ACMRed}
    \addlegendimage{line width=0.4mm,mark size=3pt,mark=pentagon,color=ACMPurple}
    \end{customlegend}
\end{tikzpicture}
\\[-\lineskip]
\vspace{-3mm}
\subfloat[Varying $\alpha$ in \algo]{
\begin{tikzpicture}[scale=1,every mark/.append style={mark size=3pt}]
    \begin{axis}[
        height=\columnwidth/2.0,
        width=\columnwidth/1.5,  
        ylabel={\it Accuracy},
        xmin=0.5, xmax=9.5,
        ymin=0.5, ymax=0.9,
        xtick={1,3,5,7,9},
        ytick={0.5,0.6,0.7,0.8,0.9},
        xticklabel style = {font=\small},
        yticklabel style = {font=\small},
        xticklabels={0.25,0.5,1,2,4,},
        yticklabels={0.5,0.6,0.7,0.8,0.9},
        every axis y label/.style={font=\small,at={(current axis.north west)},right=5mm,above=0mm},
        legend style={fill=none,font=\small,at={(0.02,0.99)},anchor=north west,draw=none},
    ]
    \addplot[line width=0.4mm, mark=triangle,color=teal]  
        plot coordinates {
(1,	0.8032	)
(3,	0.8244	)
(5,	0.8111	)
(7,	0.8003	)
(9,	0.7831	)

    };
    \addplot[line width=0.4mm, mark=diamond,color=orange]  
        plot coordinates {
(1,	0.6885 )
(3,	0.7356	)
(5,	0.7596	)
(7,	0.7279	)
(9,	0.7314	)

    };
    \addplot[line width=0.4mm, mark=square,color=ACMDarkBlue]  
        plot coordinates {
(1,	0.7643	)
(3,	0.7404	)
(5,	0.8071	)
(7,	0.8197	)
(9,	0.7915	)

    };

        \addplot[line width=0.4mm, mark=star,color=ACMRed]  
        plot coordinates {
(1,	0.7098	)
(3,	0.7258	)
(5,	0.7481	)
(7,	0.6796	)
(9,	0.5495	)

    };
        \addplot[line width=0.4mm, mark=pentagon,color=ACMPurple]  
        plot coordinates {
(1,	0.7437	)
(3,	0.7438	)
(5,	0.7600	)
(7,	0.7134	)
(9,	0.7538	)

    };
    \end{axis}
\end{tikzpicture}\hspace{5mm}\label{fig:alpha_param}%
}
\subfloat[Varying $T$ in \algo]{%
\begin{tikzpicture}[scale=1,every mark/.append style={mark size=3pt}]
    \begin{axis}[
        height=\columnwidth/2.0,
        width=\columnwidth/1.5,  
        ylabel={\it Accuracy},
        xmin=0.5, xmax=9.5,
        ymin=0.65, ymax=0.85,
        xtick={1,3,5,7,9},
        ytick={0.65,0.7,0.75,0.8,0.85},
        xticklabel style = {font=\small},
        yticklabel style = {font=\small},
        xticklabels={1,2,3,4,5},
        yticklabels={0.65,0.7,0.75,0.8,0.85},
        every axis y label/.style={font=\small,at={(current axis.north west)},right=5mm,above=0mm},
        legend style={fill=none,font=\small,at={(0.02,0.99)},anchor=north west,draw=none},
    ]
    \addplot[line width=0.4mm, mark=triangle,color=teal]  
        plot coordinates {
(1,	0.7737	)
(3,	0.8192	)
(5,	0.8212	)
(7,	0.8172	)
(9,	0.7955	)
    };
    \addplot[line width=0.4mm, mark=diamond,color=orange]  
        plot coordinates {
(1,	0.7373	)
(3,	0.7626	)
(5,	0.7463	)
(7,	0.7374	)
(9,	0.7216	)
    };
    \addplot[line width=0.4mm, mark=star,color=ACMRed]  
        plot coordinates {
(1,	0.7324	)
(3,	0.7474	)
(5,	0.7327	)
(7,	0.7425	)
(9,	0.7320	)
    };
    \addplot[line width=0.4mm, mark=pentagon,color=ACMPurple]  
        plot coordinates {
(1,	0.7425	)
(3,	0.7418	)
(5,	0.7002	)
(7,	0.7590	)
(9,	0.7069	)
    };
    \end{axis}
\end{tikzpicture}\hspace{5mm}\label{fig:t_param}}
\subfloat[Varying $\tau$ in \algo]{%
\begin{tikzpicture}[scale=1,every mark/.append style={mark size=3pt}]
    \begin{axis}[
        height=\columnwidth/2.0,
        width=\columnwidth/1.5,  
        ylabel={\it Accuracy},
        xmin=0.5, xmax=13.5,
        ymin=0.6, ymax=0.85,
        xtick={1,3,5,7,9,11,13},
        ytick={0.6,0.65,0.7,0.75,0.8,0.85},
        xticklabel style = {font=\small},
        yticklabel style = {font=\small},
        xticklabels={4,8,16,32,64,128,256},
        yticklabels={0.6,0.65,0.7,0.75,0.8,0.85},
        every axis y label/.style={font=\small,at={(current axis.north west)},right=5mm,above=0mm},
        legend style={fill=none,font=\small,at={(0.02,0.99)},anchor=north west,draw=none},
    ]
    \addplot[line width=0.4mm, mark=triangle,color=teal]  
        plot coordinates {
(1,	0.7991	)
(3,	0.7618	)
(5,	0.7862	)
(7,	0.7973	)
(9,	0.7692	)
(11, 0.8225)
(13, 0.7746)
    };
    \addplot[line width=0.4mm, mark=diamond,color=orange]  
        plot coordinates {
(1,	0.6614	)
(3,	0.7304	)
(5,	0.6904	)
(7,	0.7348	)
(9,	0.7493	)
(11, 0.7605)
(13, 0.7546)
    };
    \addplot[line width=0.4mm, mark=square,color=ACMDarkBlue]  
        plot coordinates {
(1,	0.8298	)
(3,	0.7658	)
(5,	0.7882	)
(7,	0.7550	)
(9,	0.7328	)
(11, 0.8104)
(13, 0.8370)
    };
    \addplot[line width=0.4mm, mark=star,color=ACMRed]  
        plot coordinates {
(1,	0.6839	)
(3,	0.6776	)
(5,	0.6855	)
(7,	0.6313	)
(9,	0.7184	)
(11, 0.7135)
(13, 0.7504)
    };
    \addplot[line width=0.4mm, mark=pentagon,color=ACMPurple]  
        plot coordinates {
(1,	0.7182	)
(3,	0.7425	)
(5,	0.7202	)
(7,	0.6760	)
(9,	0.7400	)
(11, 0.7516)
(13, 0.6970)
    };
    \end{axis}
\end{tikzpicture}\hspace{5mm}\label{fig:tau_param}
}
\vspace{-2mm}
\caption{Varying parameters in \algo.} \label{fig:more_param}
\vspace{-1ex}
\end{figure*}

\begin{table*}[!t]
\centering
\renewcommand{\arraystretch}{0.9}
\caption{Comparison with clustering approaches.}\vspace{-3mm}
\begin{small}
\addtolength{\tabcolsep}{-0.25em}
\resizebox{\textwidth}{!}{%
\begin{tabular}{c|c c c c|c c c c|c c c c|c c c c|c c c c}
\hline
\multirow{2}{*}{\bf Method} & \multicolumn{4}{c|}{\bf{ {\em Cora}}} & \multicolumn{4}{c|}{\bf{ {\em CiteSeer}}} & \multicolumn{4}{c|}{\bf{ {\em PubMed}}} & \multicolumn{4}{c|}{\bf{ {\em WikiCS}}}  & \multicolumn{4}{c}{\bf{ {\em DBLP}}}  \\ \cline{2-21}
& Acc & NMI & ARI & F1 & Acc & NMI & ARI & F1 & Acc & NMI & ARI & F1 & Acc & NMI & ARI & F1  & Acc & NMI & ARI & F1 \\ \hline
\texttt{AGC-DRR}~\cite{gong2022attributed} &66.97&53.66&46.14&63.21&69.56&48.54&49.42&65.97&\underline{65.87}&25.10&\underline{25.78}&65.48&57.17&46.06&39.89&48.46&74.04&\textbf{44.17}&45.98&72.66 \\ 
\texttt{AGCN}~\cite{peng2021attention}&62.88&47.79&35.82&53.86&61.45&40.79&37.15&54.97&65.09&26.39&25.59&\underline{65.93}&50.38&40.98&32.83&38.13&\underline{75.59}&43.51&\underline{49.67}&\underline{73.05} \\ 
\texttt{CCGC}~\cite{yang2023cluster}&\underline{74.62}&\underline{56.71}&\underline{53.86}&\underline{71.46}&49.45&33.23&35.11&44.29&44.16&17.74&17.68&44.14&32.08&21.06&7.66&30.76&72.85&40.92&45.02&70.61 \\ 
\texttt{DAEGC}~\cite{wang2019attributed}&72.62&53.40&49.51&70.06&69.08&47.04&47.55&65.21&62.97&22.61&21.15&63.59&\underline{59.09}&\underline{47.30}&\underline{45.72}&\underline{52.64}&75.04&43.70&48.92&72.74 \\ 
\texttt{DFCN}~\cite{tu2021deep} &70.10&50.57&44.61&66.13&71.12&\underline{48.65}&50.89&\textbf{66.46}&62.68&\underline{22.94}&21.28&63.28&52.89&43.75&30.57&40.15&74.74&43.91&48.97&72.37 \\ 
\texttt{EFR-DGC}~\cite{hao2023deep} &64.55&52.15&43.40&61.47&\underline{72.40}&48.86&50.14&\underline{66.18}&62.63&21.99&20.59&63.11&56.70&41.28&41.14&42.98&73.30&41.93&46.93&70.71 \\ 
\texttt{GCAE}~\cite{yan2021graph}&65.07&51.80&43.50&61.29&74.18&49.27&\underline{51.77}&67.00&58.15&15.25&13.94&58.02&56.19&48.68&40.61&48.24&72.15&40.59&43.77&70.39 \\ 
\hline
\algo &\textbf{81.80}&\textbf{64.22}&\textbf{64.30}&\textbf{79.64}&\textbf{76.16}&\textbf{50.87}&\textbf{54.14}&65.28&\textbf{83.70}&\textbf{48.86}&\textbf{56.56}&\textbf{83.26}&\textbf{74.52}&\textbf{54.15}&\textbf{57.37}&\textbf{69.29}&\textbf{76.37}&\underline{44.14}&\textbf{50.75}&\textbf{74.12} \\ 
Improv. & 7.18 & 7.51 & 10.44 & 8.18 & 1.98 & 1.6 & 2.37 & -1.72 & 17.83 & 22.47 & 30.78 & 17.33 & 15.43 & 5.47 & 11.65 & 16.65 & 0.78 & -0.03 & 1.08 & 1.07 \\


 \hline
\end{tabular}
}
\end{small}
\label{tbl:cluster}
\vspace{-2mm}
\end{table*}

\subsection{Datasets and Metrics}
\label{sec:dataset}
\textit{Cora}, \textit{Citeseer} and \textit{Pubmed} are three widely used datasets in the GNN community. Each node indicates a paper, and the edges indicate the citation relation. We get the raw text of each paper from \cite{chen2024exploring}. \textit{Wiki-CS} consists of nodes corresponding to Computer Science articles, with edges based on hyperlinks and 10 classes representing different branches of the field~\cite{mernyei2022wikicswikipediabasedbenchmarkgraph}.
More details can be found in Table ~\ref{tab:stats_full}. About the metrics, NMI measures the information shared between the
predicted label and the ground truth. When data are partitioned perfectly, the NMI score is 1, while it becomes 0 when prediction and ground truth are independent. ARI is a metric to measure the degree of agreement between the cluster and golden distribution, which ranges in [-1,1]. The more consistent the two distributions, the higher the score.


\subsection{Hyper-parameter Settings}
\label{sec:hyper}
We present the selection of all hyper-parameters in Table~\ref{tbl:combined_param_selection}. For all GAT-based methods, we follow the settings from~\cite{velivckovic2017graph}, where the number of attention heads is set to 8, and the number of output heads is set to 1. Specifically, for the \textit{PubMed} dataset, the number of output heads is set to 8.

For all GCNII-based methods, we adhere to the settings outlined in~\cite{chen2020simple}. In particular, for \textit{Cora}, the number of layers is set to 64, for \textit{Citeseer} the number of layers is 32, and for \textit{WikiCS} and \textit{PubMed}, the number of layers is 16.

\section{Additional Experiment Results}\label{sec:add-exp}

\subsection{More Hyper-parameter Analysis}
In this section, we further analyze the effect of $\alpha$, $T$, and $\tau$. These parameters are crucial for the initial active selection process introduced in Section~\ref{sec:initial}. We investigate the impact of each parameter and present the results in Figure~\ref{fig:more_param}. The results indicate that varying $\tau$ has the most significant effect on overall performance, particularly on the \textit{WikiCS} dataset. 

$T$ and $\alpha$ are two factors in Eq.~\eqref{eq:initial_combine} that represent the depth of the encoded representations and the importance of each hop. As shown in Figure~\ref{fig:t_param}, in most cases, setting $T=2$ and $\alpha=1$ is sufficient to achieve excellent results.

\subsection{{Comparison with Graph Clustering Methods}}

\textsf{Label-free} node classification is similar to \textsf{Clustering}, with the key distinction being the use of external category information. Clustering methods can be evaluated by mapping the predicted clustering assignment vector to the ground truth labels using the Kuhn-Munkres algorithm~\cite{plummer1986matching}. 

In Table~\ref{tbl:cluster}, we present the performance of various clustering methods across the five datasets and compare them with \algo. We highlight the best result in bold and the runner-up with an underline. The results demonstrate that \algo outperforms all clustering methods, achieving an improvement in accuracy of up to 17.83\%. These findings highlight the superiority of the proposed pipeline over traditional clustering approaches.

\begin{table}[!t]
\centering
\caption{Performance of \algo with various LLMs.}
\label{tab:gpt4}
\vspace{-2ex}
\begin{tabular}{c|c|c}
\hline
 & \textbf{\em Cora} & \textbf{\em Citeseer}  \\
\hline
\algo (GCN) with GPT-3.5-turbo& \textbf{81.80} & \textbf{76.16}  \\ 
\algo (GCN)  with GPT-4-turbo& 79.50 & 74.79 \\ 
\algo (GCN)  with GPT-4o& 80.21 & 74.87 \\ 
\hline
 \algo (GAT) with GPT-3.5-turbo& 76.06 & 73.97\\
  \algo (GAT) with GPT-4-turbo& 77.95 & 74.52\\
    \algo (GAT) with GPT-4o& 75.94 & 73.53  \\
  \hline
 \algo (GCNII) with GPT-3.5-turbo & 78.71 & 73.14\\
  \algo (GCNII) with GPT-4-turbo & 77.83 & 71.39 \\
    \algo (GCNII) with GPT-4o & 76.99 & 70.39 \\
\hline
\end{tabular}
\end{table}

\begin{table}[!t]
\centering
\caption{Accuracy of LLM-generated Annotations}
\label{tab:pseudo_quality}
\vspace{-2ex}
\begin{tabular}{c|c|c}
\hline
 \bf{ {\em Cora}} & \bf{ {\em Citeseer}} & \bf{ {\em DBLP}} \\
\hline
 73.7 & 74.0 & 75.0  \\ 
\hline
\end{tabular}
\end{table}

\begin{table}[!t]
\centering
\caption{\algo with Ground-truth Annotations}
\label{tab:gt_training}
\vspace{-2ex}
\begin{tabular}{c|c|c}
\hline
 & \algo &\algo (ground-truth)  \\
\hline
\bf{ {\em Cora}}& 81.8 & 84.7 \\ 
\bf{ {\em Citeseer}}& 76.1 & 76.2 \\ 
\bf{ {\em DBLP}}& 75.6 & 78.8 \\
\hline
\end{tabular}
\end{table}

\subsection{Evaluating  \algo with More GPT Models}

We also conducted an experiment using more advanced models, GPT-4 and GPT-4o, to annotate labels and perform the task of node classification. The results are presented in Table~\ref{tab:gpt4}. Interestingly, in some cases, \algo with noisier labels produced by GPT-3.5-turbo outperforms the higher-quality labels generated by GPT-4-turbo or GPT-4o. This indicates that, with the assistance of \algo, the performance of the LLM is no longer the sole bottleneck in zero-shot node classification. Even when using a weaker model, we can achieve results comparable to those of a stronger model, while the query costs between weaker and stronger models can differ by orders of magnitude.

\eat{
\subsection{Ablation Study on Active Node Selection}
To verify the efficacy of our subspace clustering technique for active node selection in Section~\ref{sec:selection}, we substitute $K$-Means for it in \algo. As reported in Table~\ref{tab:kmeans}, the variant of \algo with $K$-Mean consistently exhibits remarkable performance degradation compared to the version using the subspace clustering. For example, on the {\em Pubmed} dataset, the subspace clustering approach attains a large margin of $12.12\%$ in terms of classification accuracy compared to $K$-Means. The reason is that $K$-Means tends to group the majority of nodes into only a few clusters, making it hard to select sufficient representative nodes for annotation. In contrast, as elaborated in Section~\ref{sec:selection}, our subspace clustering approach can filter out the noise and identify the inherent structure underlying the node features for a more balanced partition.
}


\subsection{Ablation Study on LLM-based Annotations}
This section empirically studies the quality of LLM-generated node annotations that are used for training. 
First, on three representative datasets {\em Cora}, {\em CiteSeer}, and {\em DBLP}, the LLM-generated annotations, i.e., the pseudo-labels of nodes output by the LLM, are of an accuracy of around $75\%$, as reported in Table~\ref{tab:pseudo_quality}.
However, Table~\ref{tab:gt_training} shows that our \algo model trained using such LLM-generated pseudo-labels is able to achieve competitive classification performance when compared to that by \algo trained with the ground-truth labels. Particularly, on the {\em CiteSeer} dataset, both versions achieve a classification accuracy of $76.1\%$ and $76.2\%$, respectively. The results manifest the strong effectiveness of the self-training architecture, informative sample selection and hybrid label refinement modules in our \algo.


}

\end{document}